%% file: main.tex
\documentclass[lettersize,journal]{IEEEtran}
\usepackage{amsmath,amsfonts}
\usepackage{array}
\usepackage[caption=false,font=normalsize,labelfont=sf,textfont=sf]{subfig}
\usepackage{textcomp}
\usepackage{stfloats}
\usepackage{url}
\usepackage{verbatim}
\usepackage{graphicx}
\usepackage{cite}
\usepackage{booktabs}
\usepackage{multirow}
\usepackage{balance}
\usepackage{graphics} 
\usepackage{algorithm}
\usepackage{algpseudocode}

\usepackage{tcolorbox}  
\usepackage{graphicx}   
\usepackage{caption}    
\usepackage{enumitem}   
\usepackage{makecell}
\usepackage[font=footnotesize]{caption} 

\usepackage{xcolor}
\newcommand{\rev}[1]{\textcolor{black}{#1}}
\newcommand{\revtwo}[1]{\textcolor{black}{#1}}

\begin{document}
\title{Whole-Body Bilateral Teleoperation with Multi-Stage Object Parameter Estimation for Wheeled Humanoid Locomanipulation}

\author{
Donghoon Baek$^{1}$,
Amartya Purushottam$^{2}$,
Jason J. Choi$^{3,\dagger}$,
and Joao Ramos$^{1,2,\dagger}$
\thanks{
This work was supported by the National Science Foundation under
Grant IIS-2024775 and the UCLA Samueli School of Engineering.

$^{1}$Donghoon Baek and Joao Ramos are with the Department of
Mechanical Science and Engineering, University of Illinois
Urbana-Champaign, Urbana, IL 61801 USA. $^{2}$Amartya Purushottam and Joao Ramos are with the Department of Electrical and Computer Engineering, University of Illinois
Urbana-Champaign, Urbana, IL 61801 USA. $^{3}$Jason J. Choi is with the Department of Electrical and Computer Engineering, University of California, Los Angeles,
Los Angeles, CA 90095 USA. $^{\dagger}$Jason J. Choi and Joao Ramos contributed equally to advising this work. Donghoon Baek is the corresponding author
(e-mail: \texttt{dbaek4@illinois.edu}).
}
}
\markboth{IEEE TRANSACTIONS ON ROBOTICS, VOL. XX, NO. X, July 2026}{IEEE TRANSACTIONS ON ROBOTICS, VOL. XX, NO. X, July 2026}%

\maketitle

\begin{abstract}
This paper presents an object-aware whole-body bilateral teleoperation framework for wheeled humanoid loco-manipulation. This framework combines whole-body bilateral teleoperation with an online multi-stage object inertial parameter estimation module. The multi-stage process sequentially integrates a vision-based object size estimator, an initial parameter guess generated by a large vision-language model (VLM), and a decoupled hierarchical sampling strategy. The visual size estimate and VLM prior offer a strong initial guess of the object's inertial parameters, significantly reducing the search space for sampling-based refinement and improving the overall estimation speed. A hierarchical strategy first estimates mass and center of mass, then infers inertia from object size to ensure physically feasible parameters, while a decoupled multi-hypothesis scheme enhances robustness to VLM prior errors. Our estimator operates in parallel with high-fidelity simulation and hardware, enabling real-time online updates. The estimated parameters update the wheeled humanoid's equilibrium point and compensate for object dynamics, letting the operator focus on task execution while improving haptic force feedback, dynamic synchronization, and manipulation tracking without sacrificing compliant behavior. We validate the system on a customized wheeled humanoid with a robotic gripper and human–machine interface, demonstrating real-time execution of lifting, delivering, and releasing tasks with a payload weighing approximately one-third of the robot’s body weight.
\end{abstract}


\input{Intro}

\input{related_work}

\input{background}
\input{method}

\input{Experiment}
\input{Conclusions_FutureWork}
\input{appendix}

\section*{Acknowledgment}
The authors thank Bo Peng, Ziyi Wang, and Mayank Dubey for their assistance with the experiments. AI tools, including OpenAI ChatGPT and Anthropic Claude, were used
for language editing.

\bibliographystyle{IEEEtran} 
\bibliography{main}         
\input{biographies}

\end{document}

%% file: intro.tex
\section{Introduction}
\label{introduction}
\IEEEPARstart{H}{umanoids}, both bipedal and wheeled, are emerging as powerful platforms for physical tasks in human-centered environments. Their anthropomorphic structure enables interaction with tools and infrastructure designed for humans, making them well-suited for service, logistics, and industrial applications. Advances in hardware and control have enabled humanoids to traverse uneven terrain, run, jump, perform manipulation tasks, and execute complex motions such as backflips, demonstrating their agility and versatility in real-world settings. \cite{purushottam2022hands, purushottam2023dynamic, purushottam2024wheeled, chignoli2021humanoid, hong2022agile, sim2022tello}.

Despite significant advances, robots still fall short of human-level proficiency in physical interaction. Tasks such as lifting and transporting heavy objects remain challenging, as they require coordinated whole-body motion, force regulation under compliance, balance control, and adaptation to changing object properties and environments. In contrast, humans naturally adjust posture, modulate force, and anticipate object behavior during execution, reflecting an embodied understanding of dynamics, contact, and environmental variability. This gap raises a fundamental question: how can we leverage human motor skills to endow robots with similar adaptability and robustness?

\begin{figure}[!t]
\centering
\includegraphics[width=3in]{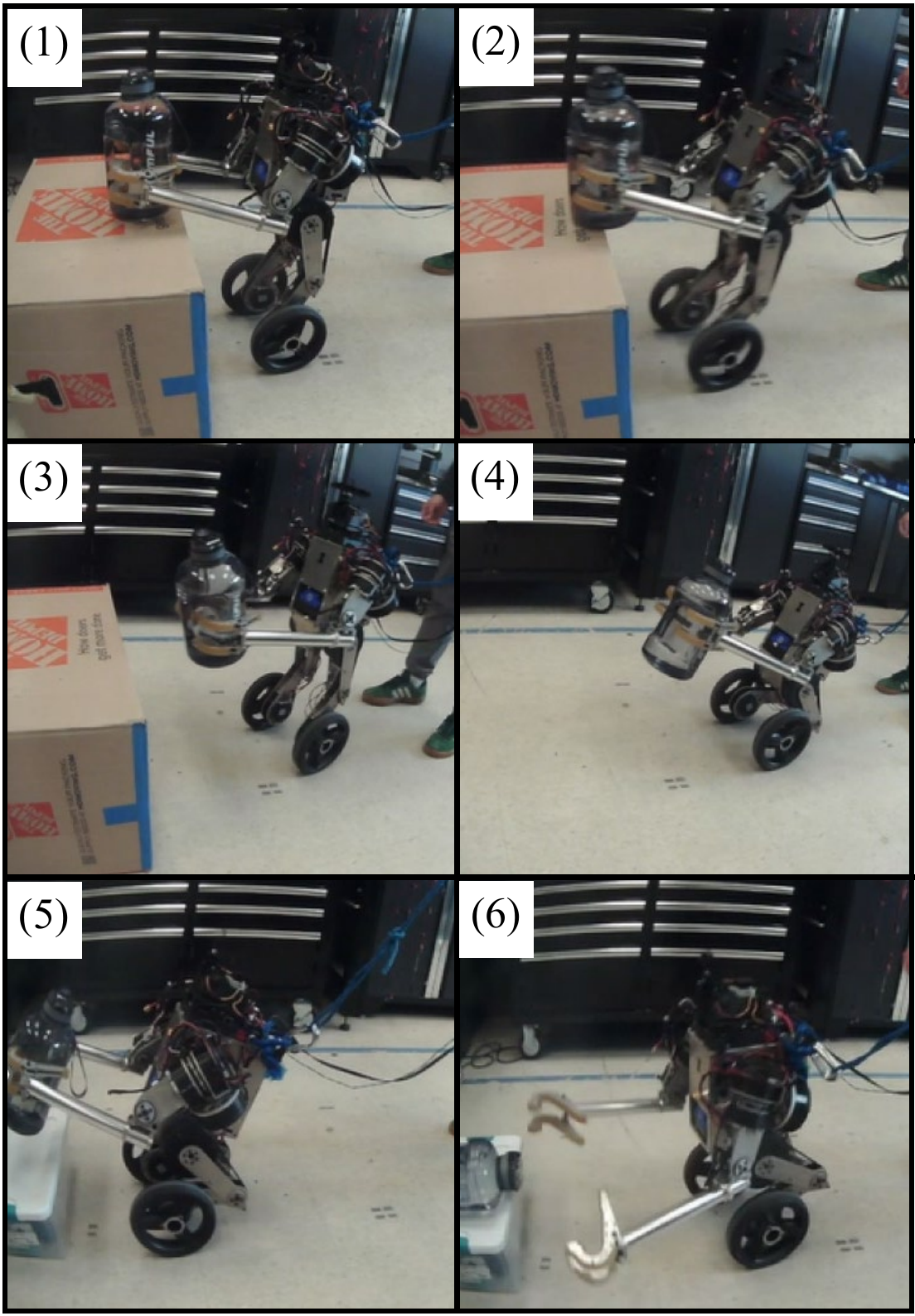}
\caption{Lifting and delivering a heavy water bottle ($\sim$1/3 of robot’s weight) using the wheeled humanoid SATYRR via whole-body bilateral teleoperation with multi-stage object parameter estimation. The framework enables SATYRR to perform squatting, lifting, and delivering motions while maintaining a stable grasp and balance under load.}
\label{fig1}
\vspace{-0.5cm}
\end{figure}

We attribute this effectiveness to two complementary factors. First, incorporating external physical information enables robots to reason about dynamic interactions and adapt their behavior, analogous to how humans leverage prior knowledge and context. Second, combining human teleoperation with robotic autonomy enables robust performance in complex environments: the human provides high-level intent, while the robot handles low-level control and physical interaction (e.g., force regulation and coordination) based on system dynamics.

Building on these hypotheses, this work develops an object-aware whole-body bilateral teleoperation framework that integrates 3D object inertial parameter estimation, enabling dynamic whole-body tele-loco-manipulation beyond human–robot synchronization. This allows the robot to adapt its equilibrium point to object dynamics, improving adaptability, accuracy, and compliant interaction, while the human focuses on high-level decision-making.

Unlike conventional loco-manipulation tasks involving lightweight objects (e.g., pick-and-place or door opening), the proposed framework enables lifting and delivering heavy objects ($\sim$1/3 of the robot’s weight) via online parameter estimation. This improves tracking accuracy while preserving compliant behavior. Fig.~\ref{fig1} illustrates a representative result.

Our proposed framework extends the concept of dynamic human–robot synchronization \cite{ramos2019dynamic,purushottam2023dynamic} by incorporating online estimation of object inertial parameters. Previous work has demonstrated the benefits of synchronizing human and robot motion using dynamic similarity and haptic force feedback. However, when external forces are large, the human pilot must manage complex decision-making, including simultaneous control of locomotion, manipulation, and disturbance rejection, often experiencing excessive force feedback. By estimating object dynamics, our approach enables the robot to update its coordination based on the new equilibrium point and compensate for the object's influence in the manipulation controller. This reduces unexpected disturbances at the interface, allowing the operator to focus more effectively on the task while maintaining meaningful and stable haptic feedback. Moreover, our object parameter estimation framework addresses three key limitations of conventional inertial parameter identification methods: 
(1) \rev{the reliance on persistently exciting trajectories for direct inertia tensor estimation, which we mitigate by leveraging a structured geometric parameterization that enables analytical computation of inertia from estimated object shape ~\cite{nadeau2022fast, zhang2025provably}}; 
(2) the dependence on force/torque sensors \cite{sundaralingam2021hand}; and 
(3) the difficulty of efficiently obtaining physically feasible inertial parameters without relying on constrained nonlinear optimization \cite{wensing2017linear,traversaro2016identification,atkeson1985rigid,lee2018geometric}. 
By eliminating these constraints, our approach enables online estimation directly on physical robot hardware during task execution.

\subsection{Contributions}
The contributions of this work are summarized as follows. First, we propose a multi-stage online estimation method for 3D object inertial parameters. Unlike most prior approaches, \rev{our method reduces reliance on long-duration persistently exciting trajectories and eliminates the need for force/torque sensors by inferring the inertia tensor from a structured geometric model rather than estimating it directly from interaction data.} The pipeline integrates (i) vision-based object size estimation, (ii) priors from a vision--language model (VLM), and (iii) a decoupled hierarchical sampling strategy to rapidly identify physically feasible parameters. Incorporating VLM priors significantly accelerates convergence by narrowing the search space, while the decoupled sampling-based refinement is executed in parallel using high-fidelity rigid-body simulation and hardware.

Second, we develop an object-aware whole-body bilateral teleoperation framework using a wheeled humanoid robot and a customized human–machine interface. This framework extends the previous Dynamic Mobile Manipulation (DMM)\cite{purushottam2023dynamic} approach to handle interactions with an unknown object—such as lifting, delivering, and releasing—without assuming that the physical properties of the manipulated object are known. Incorporating object parameter estimation enables the human operator to make more effective use of whole-body haptic force feedback while focusing freely on completing the loco-manipulation task.

The third contribution is a sim-to-real adaptation strategy designed to bridge the \textit{reality gap} for sampling-based parameter estimation using rigid-body simulation. Unlike domain randomization in reinforcement learning, which adapts via feedback from proprioception, parameter estimation lacks such a feedback loop and is thus more sensitive to sim-to-real discrepancies. By establishing a direct link between hardware and high-fidelity simulation, the proposed strategy improves the accuracy and reliability of parameter estimation.

\rev{Finally, we demonstrate an application of the proposed estimator to robust adaptive safety-critical manipulation using control barrier functions in simulation. The estimated object parameters adapt the system's dynamic constraints, and a robust formulation handles estimation uncertainty to ensure reliable control performance. Hardware validation is left as future work.}

We validate our approach through experiments with a wheeled humanoid, robotic gripper, and human–machine interface, all specifically customized for this study. The results demonstrate effectiveness in tasks such as pick-and-place, whole-body loco-manipulation for lifting and delivering heavy objects, and whole-body haptic force feedback.

\begin{figure*}[!t]
\centering
\includegraphics[width=18cm]{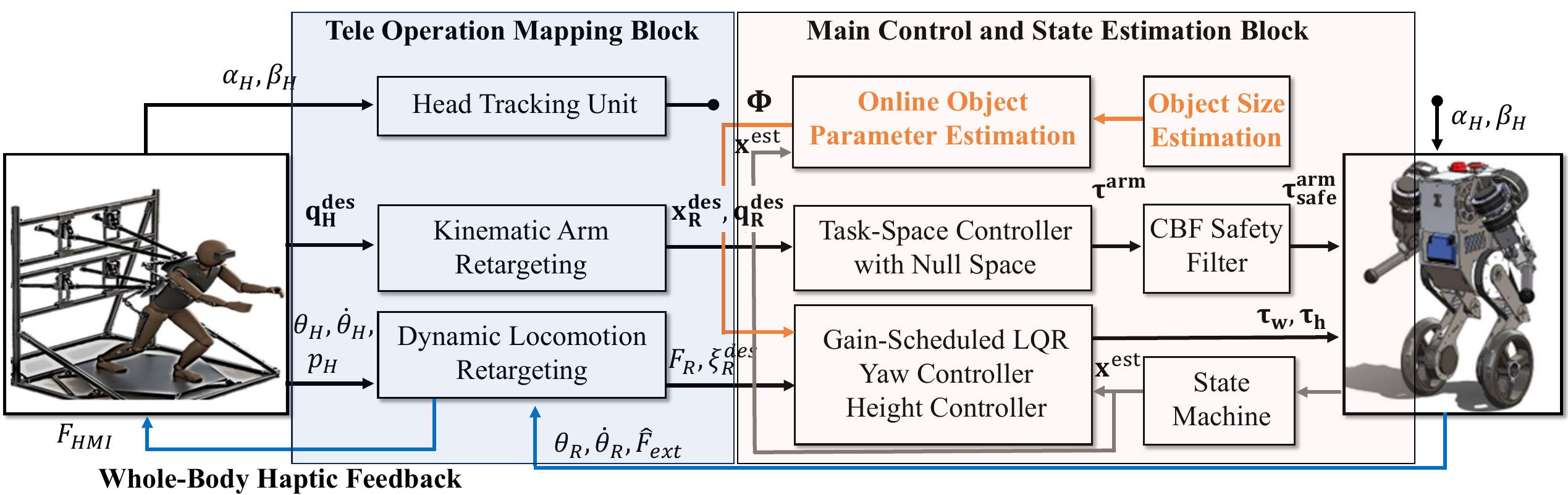}
\caption{Overview of the whole-body bilateral teleoperation framework. (Left) A human pilot controls a wheeled humanoid with haptic feedback via a human–machine interface. Wheeled locomotion is dynamically synchronized, while manipulation uses kinematic retargeting. (Right) Online multi-stage object parameter estimation runs on a rigid-body simulation, providing inertial parameters of a manipulated object to the controller. This framework enables the human to focus on high-level task execution while the robot autonomously handles low-level subtasks for efficiency and accuracy. A CBF safety filter further adapts control input for safety in autonomous setting.  
}
\label{fig2}
\end{figure*}

\subsection{System Overview and Outline}
An overview of our framework is shown in Fig. \ref{fig2}. To enable whole-body bilateral teleoperation, the human pilot sends control inputs to the robot while receiving whole-body haptic feedback through a human–machine interface (HMI). All components within the teleoperation and control modules, including the online parameter estimation, operate in real time. Specifically, the teleoperation and control loops run at 400 Hz, while the parameter estimation module runs in a separate thread at 100 Hz. 
\rev{
The vision-based size estimation runs at approximately 100 Hz, while the VLM-based prior introduces a one-time delay of 2--3 s before grasping.
The sampling-based refinement runs asynchronously in parallel with the hardware, with each update taking approximately 0.3--0.5 s. Overall, the full pipeline—from initial visual observation to stable parameter estimation—operates on a timescale of approximately 3 s in our experimental setup.
}
The parameter estimation is selectively activated—only when a transition from static to dynamic motion is detected or when a mechanical trigger is engaged. This selective update strategy avoids unnecessary computation of object inertial parameters, based on the assumption that an object's physical properties remain constant unless the manipulated object changes. Once the model is updated, it can be directly used to compensate for external torque and force. Haptic feedback is delivered through the HMI, enabling the human pilot to perceive external disturbances and intuitively adapt their motion to maintain stable and effective task execution.

%% file: related_work.tex
\section{RELATED WORK}
\label{Relatedwork}

\subsection{Control of Legged and Humanoid Robots} Over the past decades, model-based control—encompassing both reactive and predictive approaches—has advanced significantly across quadrupedal, bipedal, and wheeled humanoid platforms \cite{collins2005efficient, kajita20013d, kuindersma2016optimization, nguyen2017dynamic, raibert1986legged, tassa2012synthesis, westervelt2003hybrid}. A key advantage of these methods lies in their use of physics-based models to predict system behavior, enabling principled controller design, often formulated through optimization-based frameworks \cite{wensing2023optimization}.

In model-based control, various levels of dynamic models are utilized to capture the essential behavior of the robot. These range from reduced-order models (RoMs) such as the Wheeled Inverted Pendulum (WIP) \cite{chen2020underactuated,baek2024toward}, Linear Inverted Pendulum (LIP) \cite{colin2023whole}, single rigid-body models \cite{di2018dynamic,bledt2017policy,winkler2018gait}, and centroidal models \cite{orin2013centroidal,wensing2016improved}, to full-body dynamic models \cite{ zambella2019dynamic} that offer a more detailed and comprehensive representation of the system. While reduced-order models enable efficient planning and control by simplifying the dynamics, more complex models, such as full-body dynamics, can capture higher-order effects including joint-level coupling, distributed mass, and varying inertial properties.

Depending on the complexity of the optimal control formulation, various control architectures have been developed to balance planning horizon, model fidelity, and computational efficiency. Hierarchical approaches often separate contact planning, motion planning, and whole-body control across different time scales and model resolutions \cite{tassa2012synthesis, zhu2021contact, carius2018trajectory, wampler2009optimal, morimoto2002minimax, budhiraja2019dynamics}. In some literature, the manipulated object is modeled as an external force or disturbance (e.g., \cite{li2023multi,sleiman2021unified}), which requires continuously updating the external force or related parameters.

More recently, learning-based approaches, particularly reinforcement learning (RL), have gained prominence in legged locomotion and manipulation tasks \cite{kormushev2013reinforcement}. RL learns control policies for diverse tasks such as walking, manipulation, and navigation through interaction with the environment to learn robust and adaptive behaviors. However, this process relies heavily on high-quality data, which is typically collected in simulation due to the impracticality of extensive real-world exploration. Thanks to advances in simulation tools and sim-to-real transfer techniques, RL has demonstrated promising performance in legged robots \cite{baek2022hybrid, kormushev2013reinforcement, choi2023learning, li2023robust, lee2024integrating, baek2026bat, byrd2026adaptmanip, zhuang2023robot, fu2023deep, jenelten2024dtc}.

\subsection{Physical Parameter Estimation}
Research on inertial parameter identification of rigid bodies has primarily leveraged the linearity of system dynamics with respect to inertial parameters, enabling the use of linear least squares estimation \cite{atkeson1986estimation}. However, unconstrained optimization can yield physically inconsistent parameters. To address this, prior work has imposed physical consistency constraints through manifold optimization \cite{traversaro2016identification}, linear matrix inequalities (LMIs) \cite{wensing2017linear}, or Riemannian geometry \cite{lee2018geometric}. Recently, robust system identification has been applied to enable provably safe manipulation control \cite{zhang2025provably}. Nonetheless, these approaches are often computationally intensive, require prior shape or CAD information, and may need to be performed before task execution, potentially increasing overall task completion time.

Although force/torque (FT) sensors are commonly used for parameter estimation, their practical limitations—cost, weight, and noise—motivate sensor-free alternatives. Recent methods have estimated partial parameters using encoder feedback and attention mechanisms \cite{lao2023learning}, but full parameter recovery without torque sensing remains unsolved. 

Learning-based methods using visual data \cite{standley2017image2mass, wu2015galileo} have also been explored, but visual input alone lacks dynamic observability due to unknown density distributions. With recent advances in reinforcement learning and sim-to-real transfer, learning-based approaches have been applied to inertial parameter estimation \cite{allevato2020tunenet,chebotar2019closing,ramos2019bayessim,muratore2022neural,tsai2021droid,tiboni2023dropo,du2021auto,gruner2024pseudo,yu2017preparing}. However, they typically require multiple iterations and task-relevant training data, limiting their applicability in real-time scenarios. Acquiring such data is particularly difficult for object-level inertial parameters, which are not directly observable. A recent learning-based method \cite{baek2024online} estimates the inertial parameters of 3D objects using only robot proprioception. While it enables fast inference, its scalability across diverse motions and object geometries remains limited.

\subsection{Whole-Body Tele-Operation}
Significant efforts in the robotics community have focused on humanoid robots, teleoperation, and shared autonomy \cite{darvish2023teleoperation,baek2023study}. Key technical challenges include the choice of motion capture system, the motion retargeting strategy between human and robot, robust low-level controller design, and the level of human authority in the control loop. Various motion capture systems have been explored, including human motion capture suits, exoskeletons, haptic feedback devices, and VR-based visual feedback systems \cite{darvish2019whole, wang2021dynamic, ramos2019dynamic, penco2019multimode}. While model-based control using reduced-order models remains common \cite{chi2024novel, purushottam2023dynamic, ramos2019dynamic, ramos2018humanoid, purushottam2022hands}, Learning-based methods have recently shown strong performance. Enabled by imitation learning and sim-to-real transfer, robots have demonstrated diverse skills such as coordinated dancing, object transport, trash disposal, and basic household manipulation \cite{fu2024humanplus, lu2024mobile, xu2025intermimic, he2025asap, li2025amo, he2024learning, fu2024mobile, zhang2025falcon}. However, in wheeled humanoid systems, dynamic tasks such as lifting heavy boxes or water containers remain underexplored compared to dexterous whole-body manipulation, as they are more sensitive to environmental dynamics that can no longer be neglected.
 

\subsection{Sim-to-Real Transfer and Simulation}

With recent advances in reinforcement learning, sim-to-real transfer has become increasingly important. Techniques such as domain randomization, simulation parameter optimization, and parameter distribution reduction~\cite{chebotar2019closing, tiboni2023dropo, du2021auto, tsai2021droid, allevato2020tunenet, muratore2022neural, gruner2024pseudo} have been proposed to reduce the reality gap. However, they typically require many iterations and task-relevant data, which is difficult to acquire. In control, methods like domain randomization and privileged learning~\cite{li2023robust, hwangbo2019learning} enable zero-shot transfer but often fall short in matching state trajectories, limiting parameter estimation accuracy~\cite{li2023robust}. More recently, He et al.~\cite{he2025asap} proposed a residual action model that compensates for dynamics mismatch through corrective actions, bypassing explicit system identification. While most works focus on reducing the reality gap to improve the robustness and generalization of reinforcement learning policies, our work centers on accurate state estimation where even small mismatches can significantly degrade estimation quality. Unlike policy learning, which can tolerate some model error through corrective feedback, parameter estimation requires precise modeling to ensure consistency between simulated and real-world state trajectories.


%% file: background.tex
\section{Background}
\label{Background}

\subsection{Inertial Parameter Estimation}

The inertial parameters of the \( i \)th rigid body (e.g., a robot link) include its mass \( m_i \), the first mass moment \( \mathbf{h}_i = m_i \mathbf{p}_i \in \mathbb{R}^3 \), where \( \mathbf{p}_i \) is the center of mass (CoM), and the rotational inertia tensor \( \mathbf{I}_i \in \mathbb{R}^{3 \times 3} \), expressed at the body frame origin. The complete inertial parameter vector \( \boldsymbol{\phi}_i \in \mathbb{R}^{10} \) is given by:
\begin{equation}
    \boldsymbol{\phi}_i = [m_i, \mathbf{h}_i^\top, I_{xx}^i, I_{yy}^i, I_{zz}^i, I_{xy}^i, I_{yz}^i, I_{zx}^i]^\top.
\end{equation}

\noindent For an \( n \)-link articulated system (e.g., a humanoid robot), the equations of motion under contact are given by:
\begin{equation}
    \mathbf{M}(\mathbf{q}, \boldsymbol{\Phi})\ddot{\mathbf{q}} + \mathbf{b}(\mathbf{q}, \dot{\mathbf{q}}, \boldsymbol{\Phi}) = \boldsymbol{\tau} + \mathbf{J}_c^\top(\mathbf{q})\mathbf{f}_c,
\end{equation}
where \( \mathbf{M} \) is the configuration-dependent mass matrix, \( \mathbf{b} \) includes Coriolis, centrifugal, and gravitational terms, and \( \boldsymbol{\tau} \in \mathbb{R}^n \) denotes joint torques. The term \( \mathbf{J}_c(\mathbf{q}) \in \mathbb{R}^{k \times n} \) is the contact Jacobian, and \( \mathbf{f}_c \in \mathbb{R}^k \) represents external contact forces. The full inertial parameter vector is \( \boldsymbol{\Phi} = [\boldsymbol{\phi}_1^\top, \dots, \boldsymbol{\phi}_n^\top]^\top \in \mathbb{R}^{10n} \). 

Rigid-body dynamics can also be expressed in a linear-in-parameters form, with equations of motion as a linear combination of known functions and unknown inertial parameters:
\begin{equation}
    \boldsymbol{\tau} = \Gamma(\mathbf{q}, \dot{\mathbf{q}}, \ddot{\mathbf{q}})\boldsymbol{\Phi},
\end{equation}
where \( \Gamma \in \mathbb{R}^{n \times 10n} \) is the regressor matrix \cite{atkeson1986estimation}. This form enables efficient system identification and control design by reducing dynamics to a linear system \( \mathbf{A}\boldsymbol{\Phi} = \boldsymbol{\tau} \), which can be solved via least squares. 

In estimating the inertial parameters of an unknown object modeled as a single rigid body, a classical linear regression model based on the Newton-Euler equations is commonly used. Specifically, given the object's linear acceleration \( \mathbf{a} \), angular velocity \( \boldsymbol{\omega} \), and angular acceleration \( \dot{\boldsymbol{\omega}} \), the relationship is:
\begin{equation}
    \begin{bmatrix}
        \mathbf{f} \\
        \boldsymbol{\tau}
    \end{bmatrix}
    =
    \begin{bmatrix}
        m\mathbf{a} \\
        [\boldsymbol{\omega}]\mathbf{I} \boldsymbol{\omega} + \mathbf{I} \dot{\boldsymbol{\omega}}
    \end{bmatrix}
    = \mathbf{Y} \boldsymbol{\phi},
    \label{eqn_newton_euler_1}
\end{equation}
where \( \mathbf{f}, \boldsymbol{\tau} \in \mathbb{R}^3 \) are the force and torque vectors, \( m \in \mathbb{R}^+ \) is the mass, and \( \mathbf{I} \in \mathbb{R}^{3 \times 3} \) is the rotational inertia tensor. The notation \([ \cdot ]\) represents the skew-symmetric matrix form of a vector. All quantities are expressed in the object's body frame, unlike the robot system identification setting discussed earlier. In our case, we assume that the object frame coincides with the end-effector frame.

To obtain physically consistent inertial parameters, a constrained least squares problem over a manifold is formulated:

\begin{equation}
\begin{aligned}
\min_{\mathbf{R}, \mathbf{J}, \mathbf{c}, m} \quad & \sum_{m} \left\| \mathbf{Y}^{(m)} \, \boldsymbol{\pi}(\mathbf{R}, \mathbf{J}, \mathbf{c}, m) - \boldsymbol{\tau}^{(m)} \right\|^2, \\
\text{s.t.} \quad & \mathbf{R} \in \mathrm{SO}(3), \quad m > 0, \quad J_i > 0,\quad i = 1,2,3 \\
&  J_1 + J_2 + J_3 \ge 2 J_k, \quad \forall k \in \{1, 2, 3\}
\end{aligned}
\end{equation}

\noindent here, \( \mathbf{Y}^{(m)} \) is the regressor matrix, \( \boldsymbol{\tau}^{(m)} \) is the corresponding measured torque, and \( \boldsymbol{\pi}(\mathbf{R}, \mathbf{J}, \mathbf{c}, m) \) denotes the inertial parameter vector, defined by the rotation matrix \( \mathbf{R} \in \mathrm{SO}(3) \), the diagonal inertia matrix \( \mathbf{J} = \mathrm{diag}(J_1, J_2, J_3) \) with \( \mathbf{I} = \mathbf{R} \mathbf{J} \mathbf{R}^\top \), the center of mass \( \mathbf{c} \), and the mass \( m \). We refer readers to \cite{atkeson1986estimation,wensing2017linear,traversaro2016identification} for a detailed discussion of classical approaches to inertial parameter estimation. Due to the non-convexity of the problem, this formulation typically requires custom manifold solvers and does not guarantee global optimality \cite{traversaro2016identification}. This formulation can be extended to unconstrained optimization \cite{rucker2022smooth} and is used as our baseline.  

\subsection{Cart-Pole Model as a Simplified Representation of the SATYRR Wheeled Humanoid}

The linearized dynamics of the cart-pole system around the equilibrium point, assuming small-angle and low-velocity approximations, are expressed in state-space form:
\renewcommand{\arraystretch}{0.9}
\begin{equation}
\label{linStateSpaceModel}
\begin{bmatrix}
\dot{x}_w\\ \ddot{x}_w\\ \dot{\theta}_R\\ \ddot{\theta}_R
\end{bmatrix} =
\underbrace{\begin{bmatrix}
0 & 1 & 0 & 0\\
0 & 0 & -\frac{m}{M}g & 0\\
0 & 0 & 0 & 1\\
0 & 0 & \frac{mg}{Mh} & 0
\end{bmatrix}}_{\small A}
\begin{bmatrix}
x_w\\ \dot{x}_w\\ \theta_R\\ \dot{\theta}_R
\end{bmatrix} +
\underbrace{\begin{bmatrix}
0 \\ \frac{1}{M}\\ 0\\ -\frac{1}{M h}
\end{bmatrix}}_{\small B} u +
\underbrace{\begin{bmatrix}
0 \\ 0\\ 0\\ \frac{1}{mh}
\end{bmatrix}}_{\small B_{ext}} F,
\end{equation}
\renewcommand{\arraystretch}{1.0}

\noindent here, $x_w$ denotes the cart position, $\theta_R$ the pendulum angle, $h$ the pendulum height, $m$ the pendulum mass, $M$ the base (cart) mass, $g$ gravity, $u$ the control input, and $F$ an external force applied at the pendulum’s center of mass. 

Assuming $m \gg M$, the system satisfies $\ddot{x}_w^{\mathrm{des}} \propto \theta_R^{\mathrm{des}}$, indicating that pitch modulation allows the human pilot to regulate translational interaction forces \cite{purushottam2022hands,purushottam2023dynamic}.

\subsection{Divergent Component of Motion (DCM)}
The Divergent Component of Motion (DCM) is a key concept in modeling balance and stability for both humans and bipedal robots. It captures the unstable dynamics of the Linear Inverted Pendulum (LIP) by isolating the forward-diverging component of the center of mass (CoM) motion. Physically, the DCM represents the extrapolated position of the CoM, incorporating both position and velocity to predict where the system is heading if no corrective action is taken.

For a pendulum with CoM position \( x_o \), velocity \( \dot{x}_o \), and height \( h \), the natural frequency is \( \omega_o = \sqrt{g/h} \), and the DCM is defined as:
\[
\xi = x_o + \frac{\dot{x}_o}{\omega_o}.
\]
Stability is achieved by ensuring that \( \xi \) remains within the foot support region \([p_x^{\min}, p_x^{\max}]\), which leads to the condition:
\[
p_x^{\min} \leq \xi \leq p_x^{\max}.
\]

When linearized using the pendulum angle \( \theta_o = x_o / h \), this becomes:
\[
\frac{p_x^{\min}}{h} \leq \theta_o + \frac{\dot{\theta}_o}{\omega_o} \leq \frac{p_x^{\max}}{h},
\]
where the dimensionless DCM is defined as \( \xi = \theta_o + \frac{\dot{\theta}_o}{\omega_o} \).

This representation is widely used in balance control and motion planning because it captures the system's dynamic behavior. By actively regulating \( \xi \) through control inputs such as base motion or body posture, a robot can maintain dynamic stability during locomotion~\cite{dynStCond}.

%% file: method.tex
\section{Multi-Stage Object Physical Parameter Estimation}
\label{method}

\begin{figure}[t]
\centering
\includegraphics[width=\linewidth]{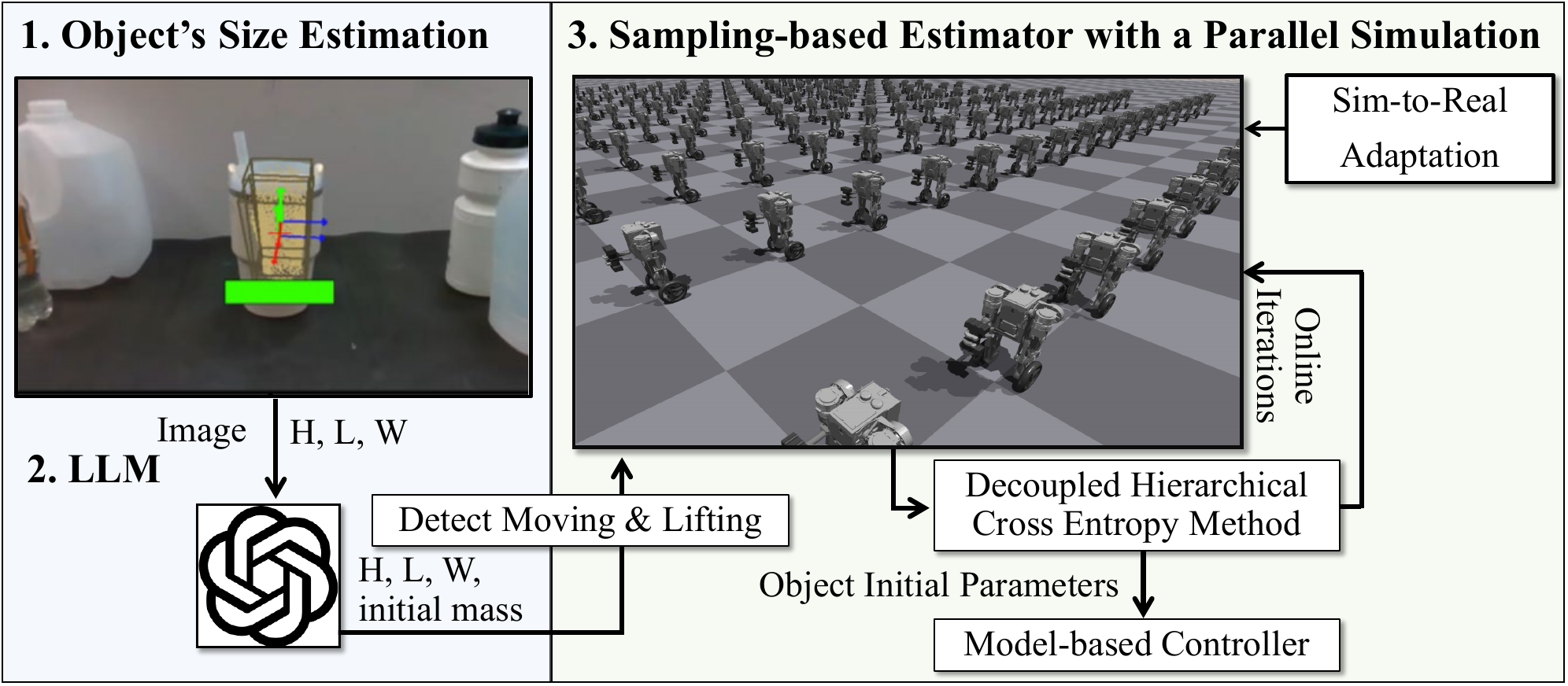}
\caption{
\rev{
Overview of the multi-stage object inertial parameter estimation pipeline. 
The framework integrates multi-modal information—vision, language instructions, and proprioception—to estimate object inertial parameters. 
(1) A vision-based module estimates object size (height, length, width) from RGB-D observations. 
(2) A vision-language model (LLM) provides an initial estimate of object mass and physical properties, triggered during object motion (e.g., lifting). 
(3) A sampling-based estimator using a decoupled hierarchical Cross-Entropy Method (DH-CEM) runs in parallel with high-fidelity simulation with sim-to-real adaptation. 
The parameters are iteratively refined and integrated into a model-based controller for locomotion and manipulation.
}
}
\label{fig_dh_cem}
\end{figure}

We present online multi-stage approach for estimating the inertial parameters of unknown 3D objects within a whole-body bilateral teleoperation framework. \rev{Unlike traditional least-squares approaches that require persistently exciting 
trajectories or force/torque sensors, \revtwo{the priors and structural decomposition} reduces this dependence 
by inferring the inertia tensor from geometric assumptions rather than 
estimating it directly from interaction data — effectively reducing the number 
of weakly observable parameter directions at the cost of inertia accuracy 
for non-cuboidal objects.} As shown in Figure. \ref{fig_dh_cem}, the proposed method combines a multi-stage sampling-based strategy with three key components: (1) prior knowledge from a vision system and vision-language model (VLM) to generate informed initial guesses and reduce the parameter search space, (2) parallelized sampling in a high-fidelity simulator, synchronized with real-world hardware, to produce diverse and realistic samples in real time using multiple robot agents, and (3) high-fidelity rigid-body simulation developed with sim-to-real adaptation offline. 

Prior to lifting, the robot obtains an initial guess of the object's inertial parameters using vision and a VLM. These estimated parameters are refined through a sampling-based method that leverages high-fidelity rigid-body simulation. By comparing the proprioceptive history between the physical hardware and simulated robots, the method identifies the simulated object whose interaction dynamics most closely match the real-world scenario. This correspondence indicates that the simulated robot is holding an object with similar physical properties. As a result, the method rapidly converges to physically consistent parameters, enabling the controller to adaptively interact with novel environments.

\subsection{Stage One: Object Size Estimation Using Vision-Based Inference}

\begin{figure}[t]
\centering
\includegraphics[width=3.5in]{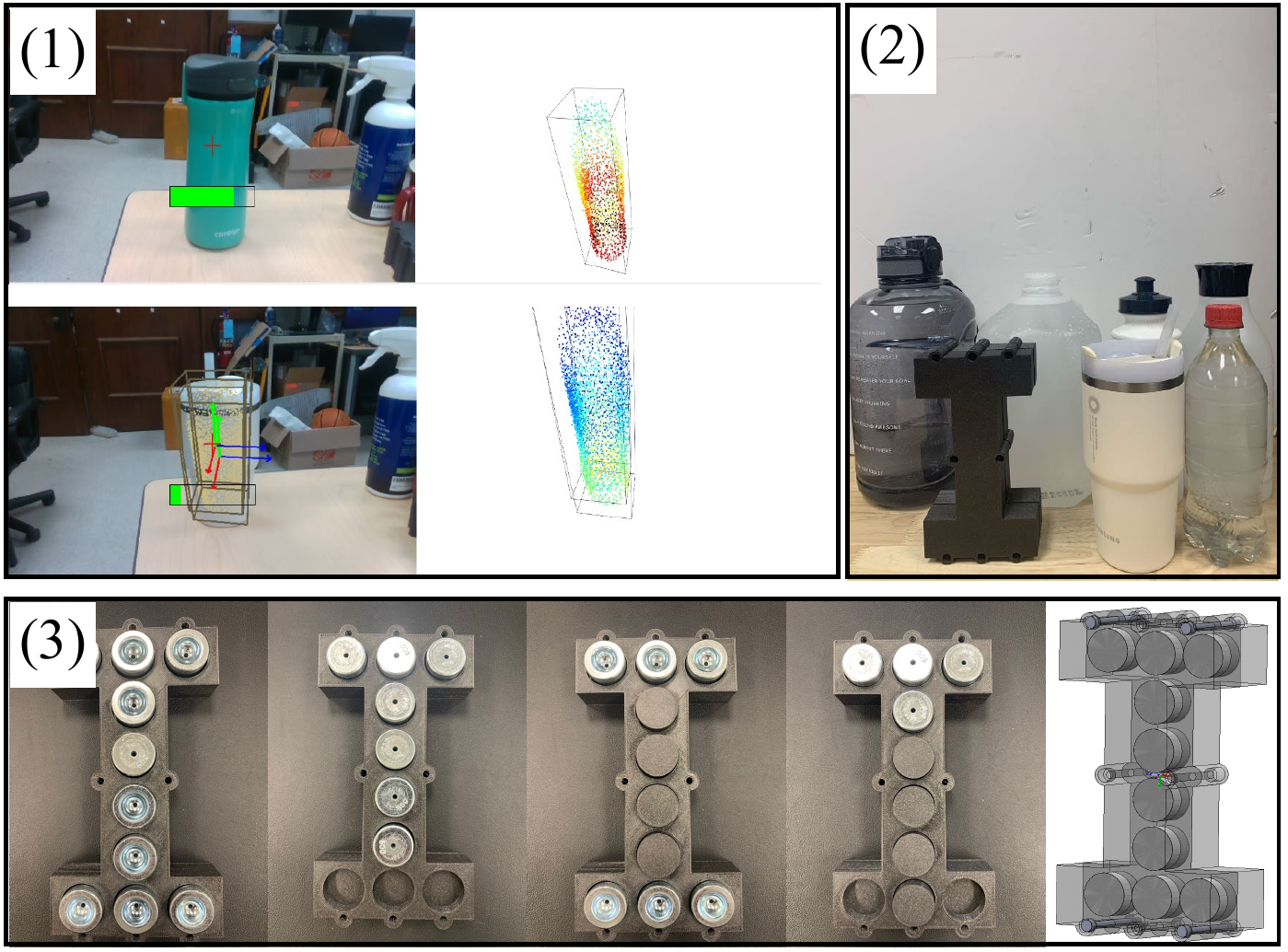}
\caption{(1) Object size estimation using CenterSnap by aligning the camera with the object center. (2) Bottle-like objects used to evaluate the proposed multi-stage object inertial parameter estimation. (3) Modular I-shaped object representing various mass distributions (e.g., full-steel, hammer, barbell) by changing internal weights.}

\label{fig3_cetersnap}
\end{figure}

We first estimate the size (length, width, and height) of a 3D object by approximating it as a cuboid using CenterSnap~\cite{irshad2022centersnap} (Fig.~\ref{fig3_cetersnap}-1). CenterSnap is a real-time (40 FPS), single-shot method that predicts the 3D shape, 6D pose, and size of objects from a single RGB-D image without requiring CAD models at test time. The method detects objects as 2D centers via a Gaussian heatmap,
\[
Y_{xy} = \exp\left(-\frac{(x - c_x)^2 + (y - c_y)^2}{2\sigma^2}\right),
\]
and predicts a 141-dimensional vector 
\[
\mathbf{o} = [z \in \mathbb{R}^{128},\, \widehat{R} \in SO(3),\, \hat{t} \in \mathbb{R}^3,\, \hat{s} \in \mathbb{R}^3],
\]
which encodes a latent shape representation, pose, and metric scale. The latent code \( z \) is decoded into a canonical point cloud \( \widehat{P} \) using a PointNet-based autoencoder trained with the Chamfer distance, and transformed into world coordinates as \( \widehat{P}_{\text{recon}} = \widehat{R}(\hat{s} \cdot \widehat{P}) + \hat{t} \). The object size is then obtained by computing the axis-aligned bounding box (AABB) of the scaled point cloud, i.e., \( \text{Size} = \max(\hat{s} \cdot \widehat{P}) - \min(\hat{s} \cdot \widehat{P}) \), yielding physical dimensions in meters (Table~\ref{tab:size_estimation_compact}).

The estimated size $\mathbf{s} = [a, b, c]^\top$ provides a strong geometric prior for subsequent estimation by constraining the center of mass (CoM) to lie within the object volume, thereby reducing the search space from an unbounded region to a bounded cuboid and significantly improving sampling efficiency.

\rev{Alternative RGB-D-based pipelines can also be used for size estimation. In our preliminary experiments, SAM2-based segmentation improved geometric consistency for irregular shapes, while YOLO-based methods showed sensitivity to detection errors on unseen objects. Based on these observations, we use CenterSnap in our final experiments due to its real-time performance and stable estimation in structured scenarios.}

\input{table/centersnap_table}

\subsection{Stage Two: Initializing the inertial parameters using a Vision Language Model}

Ideally, if the mass density $\rho$ and the volume $V$ of a rigid object are known, the full inertial parameters can be computed analytically. Specifically, the total mass is given by:
\begin{equation}
    m = \rho V,
\end{equation}
where $\rho$ [kg/m$^3$] denotes the object's (possibly uniform) density and $V$ [m$^3$] is its geometric volume. The CoM $\mathbf{h} \in \mathbb{R}^3$ and the inertia tensor $\mathbf{I} \in \mathbb{R}^{3 \times 3}$ with respect to the CoM are computed as:
\begin{align}
    \mathbf{h} &= \frac{1}{m} \int_V \mathbf{r} \rho(\mathbf{r})\, dV, \label{eq:com_integral} \\
    \mathbf{I} &= \int_V \rho(\mathbf{r}) \left( \|\mathbf{r} - \mathbf{h}\|^2 \mathbf{I}_3 - (\mathbf{r} - \mathbf{h})(\mathbf{r} - \mathbf{h})^\top \right) dV, \label{eq:inertia_integral}
\end{align}
where $\mathbf{r} \in \mathbb{R}^3$ denotes the position of a differential volume element, $\rho(\mathbf{r})$ is the spatially varying density (which reduces to constant $\rho$ for uniform-density objects), and $\mathbf{I}_3$ is the $3\times 3$ identity matrix. The term $\|\mathbf{r} - \mathbf{h}\|^2 \mathbf{I}_3$ accounts for the squared distance from the CoM (scaled across all axes), while the outer product $(\mathbf{r} - \mathbf{h})(\mathbf{r} - \mathbf{h})^\top$ encodes directional offsets used in the parallel axis theorem. These expressions provide a foundational link between geometry (object size and shape), density (material), and inertial parameters, and are particularly useful for prior-informed estimation using perceptual cues such as size and material.

Humans intuitively estimate object weight and dynamics by leveraging prior experience and perceptual cues such as size, shape, and material. Inspired by this, we utilize a Vision Language Model (VLM)~\cite{ahn2022can} to infer physically relevant priors such as approximate density, shape category, and volume from visual and textual inputs. These priors are used to initialize and constrain the sampling distribution for inertial parameter estimation, enabling more efficient convergence by focusing the search on physically plausible regions of the parameter space. 

We summarize the mass estimation results from the VLM in Table \ref{tab:percentage_improvement} and present the prompting strategy in Fig. \ref{fig:multi_step_prompt}. The estimated mass reflects the model’s reasoning over material densities (e.g., water, plastic), whether the object is filled with liquid, and the estimated fill level. We approximate the object as a uniform-density cuboid. Under this assumption, the inertia tensor of a cuboid with mass \( m_i \) and dimensions \( a_i, b_i, c_i \), offset by \( (h_x, h_y, h_z) \) from its centroid, is computed as:
\begin{equation}
\begin{aligned}
I_{xx}^i &= \tfrac{1}{12} m_i (b_i^2 + c_i^2) + m_i(h_y^2 + h_z^2), \\
I_{yy}^i &= \tfrac{1}{12} m_i (a_i^2 + c_i^2) + m_i(h_x^2 + h_z^2), \\
I_{zz}^i &= \tfrac{1}{12} m_i (a_i^2 + b_i^2) + m_i(h_x^2 + h_y^2)
\end{aligned}
\label{eq:diag_inertia_tensor}
\end{equation}

\begin{equation}
I_{xy}^i = -m_i h_x h_y,\quad
I_{yz}^i = -m_i h_y h_z,\quad
I_{zx}^i = -m_i h_z h_x.
\label{eq:offdiag_inertia_tensor}
\end{equation}

\noindent This formulation yields \( N \) full inertial parameter samples \( \{ \boldsymbol{\phi}_i \}_{i=1}^N \), which are used in the next phase of estimation.

\begin{figure}[t]
\centering
\begin{minipage}{0.95\linewidth}
\begin{tcolorbox}[
  colback=gray!5!white,
  colframe=black!75!black,
  title=Prompt for Inertial Parameter Estimation,
  fonttitle=\bfseries,
  fontupper= \scriptsize\ttfamily  
]
\textbf{Instructions}\\[0.5ex]
Your task is to estimate the inertial parameters (mass, center of mass, and inertia tensor) of a 3D rigid body object (e.g., a bottle). The input is an image of the environment containing the target object and surrounding items (e.g., containers), overlaid with the estimated dimensions of the object assuming a cuboid shape, along with a text instruction.\\

Based on the image, assess whether the object is filled with liquid (e.g., water) and whether there are any additional objects attached to or surrounding the target object.\\

To determine the inertial parameters, consider the object's estimated size, its fill level, and its material density. Assume the object can be approximated as a cuboid. If the object appears to be a familiar item, use known specifications such as its volume and weight to improve the estimation.
\end{tcolorbox}
\end{minipage}
\caption{Prompt used for visual-based inertial parameter estimation of 3D rigid objects. The instructions guide the model to infer mass, center of mass, and inertia tensor from object geometry, contents, and contextual cues in the image.}
\label{fig:multi_step_prompt}
\end{figure}

\subsection{Stage Three: Decoupled Hierarchical Cross-Entropy Method with High-Fidelity Parallel Simulation}

The inertial parameter samples obtained from vision and a VLM provide initial guesses for the Stage Three sampling-based estimator, which runs in parallel with physical hardware using a high-fidelity rigid-body simulation. To obtain more realistic samples, we apply offline sim-to-real adaptation (Appendix~\ref{appendix::sim-to-real}) to minimize the reality gap. Building on the adapted simulation, we implement a decoupled hierarchical Cross-Entropy Method (DH-CEM).

\begin{algorithm}[t]
\caption{Decoupled Hierarchical Cross Entropy Method for Inertial Parameter Estimation}
\label{alg:dhcem}
\begin{algorithmic}[1]  

\State \textbf{Phase 1: Multi-Hypothesis Initialization}
\State \textbf{Input:} Estimated mass $\hat{m}$, object size $\mathbf{s} = [a, b, c]$
\State \textbf{Initialize:} $K$ hypotheses $\{\phi\}_{i=1}^K$
\For{$i = 1$ to $K$}
    \State Sample $m_i \sim \mathcal{U}((1 - \delta_m)\hat{m},\, (1 + \delta_m)\hat{m}))$
    \State Sample $\mathbf{s}_i \sim \mathcal{U}((1 - \delta_s)\mathbf{s}, (1 + \delta_s)\mathbf{s})$
    \State Sample CoM $\mathbf{c}_i \sim \mathcal{U}(-\delta_h\mathbf{s}_i, \delta_h\mathbf{s}_i)$
    \State Compute inertia via PAT using $m_i$, $\mathbf{s}_i$, $\mathbf{c}_i$
    \State Form hypothesis $\phi_i = [m_i, \mathbf{c}_i, \text{Inertia}(m_i, \mathbf{s}_i, \mathbf{c}_i)]$
\EndFor
\State Construct initial distribution $\mathcal{N}(\mu^{(0)}, \Sigma^{(0)})$ from $\{\phi_i\}_{i=1}^K$

\vspace{0.5em}
\State \textbf{Phase 2: DH-CEM-Based Refinement in Simulation}
\For{$t = 0$ to $T_{\text{max}}$}
    \For{$j = 1$ to $N$}
        \State Sample $\phi_j^{(t)} \sim \mathcal{N}(\mu^{(t)}, \Sigma^{(t)})$
        \State Inject $\phi_j^{(t)}$ into simulation $\hat{f}_{\text{sim}}$
        \State Simulate trajectories $\{\hat{q}_j(t), \hat{\dot{q}}_j(t)\}$
        \State Compute cost: $J(\phi_j^{(t)})$
        \Statex \hspace{1.5em} $ = \sum_k \left\| \hat{\mathbf{q}}_j(k) - \mathbf{q}^*(k) \right\|^2+\left\| \hat{\dot{\mathbf{q}}}_j(k) - \dot{\mathbf{q}}^*(k) \right\|^2$
    \EndFor
    \State Select top-$M$ elite samples $\mathcal{E}^{(t)}$
    \State Update mean: $\mu^{(t+1)} = \frac{1}{M} \sum_{\phi \in \mathcal{E}^{(t)}} \phi$
    \State Update covariance: $\Sigma^{(t+1)} = \text{Cov}(\mathcal{E}^{(t)})$
\EndFor

\vspace{0.5em}
\State \textbf{Return:} Final estimate $\widehat{\phi} = \arg\min_{\phi_j^{(T)}} J(\phi_j^{(T)})$

\end{algorithmic}
\end{algorithm}


Starting from the VLM-based initial guess, DH-CEM first refines mass and center of mass from physical interactions, then infers the inertia tensor from the estimated object size and continued interaction. This hierarchical approach improves efficiency and physical plausibility compared to classical methods that estimate all parameters simultaneously.

\begin{figure}[t]
    \centering
    \includegraphics[width=0.8\linewidth]{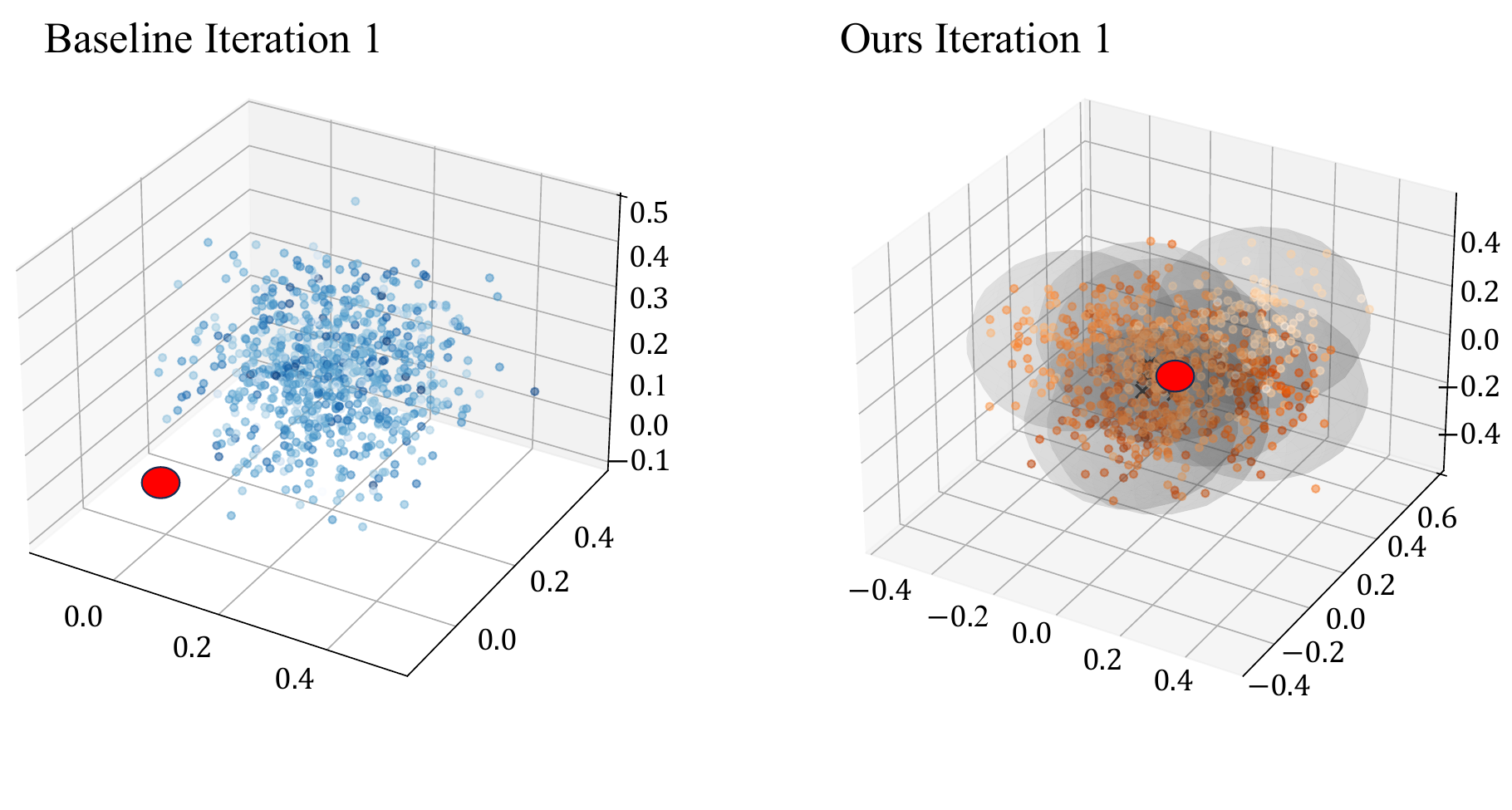} \\
    \vspace{2mm}
    \includegraphics[width=0.8\linewidth]{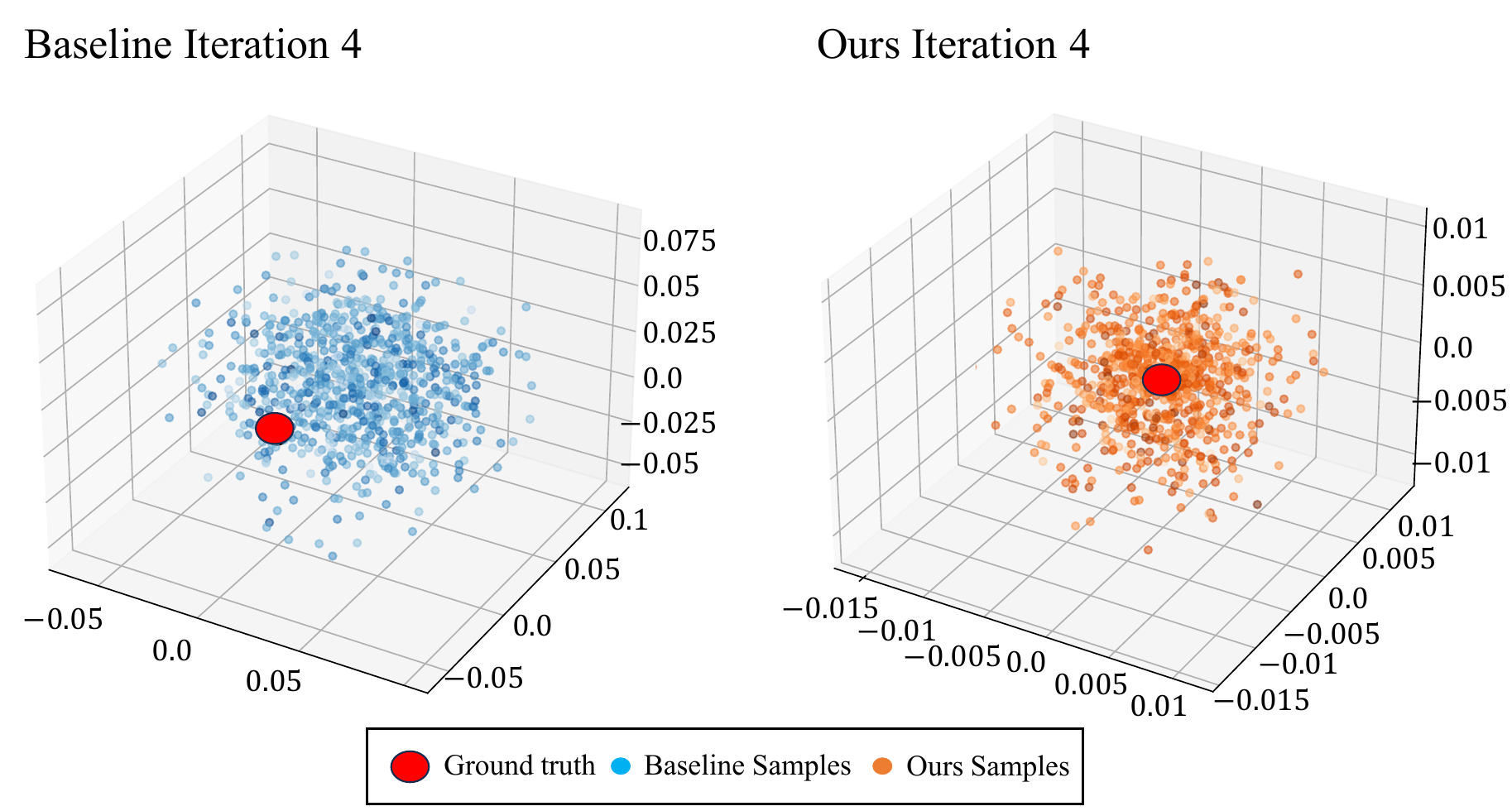}
    \caption{Sampled inertial parameter distributions at early (top) and later (bottom) iterations. The red cross indicates ground truth. This indicates that using multiple hypotheses makes the sampling-based approach less sensitive to poor initial guesses.}
    \label{fig:cem_iterations}
\vspace{-2mm} 
\end{figure}

To balance convergence speed and algorithmic simplicity, our sampling-based approach builds on the Cross-Entropy Method (CEM)—a stochastic, derivative-free optimization algorithm widely used in control and reinforcement learning. CEM iteratively updates the sampling distribution to increase the likelihood of generating top-performing trajectories. Compared to classical CEM, our approach introduces three key differences. First, we adopt a multi-hypothesis initialization strategy in which an ensemble of CEM instances is generated to capture the uncertainty associated with the vision-based size estimation and the outputs of the VLM (see Fig.~\ref{fig:cem_iterations}). For example, if the bottle is nontransparent, the VLM alone may fail to infer the material or estimate the fill level accurately. This formulation ensures better coverage of plausible hypotheses by avoiding mode collapse through broader exploration of the parameter space during the early stages of the estimation process. \revtwo{Second, our method adopts a decoupled-hierarchical structure in which mass and CoM are estimated from physical interaction, while inertia is computed analytically from the estimated object size and CoM, ensuring physically plausible parameters. This decomposition, together with the vision/VLM priors, is what reduces the number of quantities that must be identified from persistently exciting motion — the CEM-based search itself only determines how efficiently the reduced (mass, CoM) space is explored.} Third, to estimate the inertial parameters of a 3D object using only the robot's proprioceptive data, we leverage a high-fidelity rigid-body simulator (e.g., IsaacGym) to generate realistic dynamics samples. The underlying assumption is that, if the simulator accurately models real-world physics, then a simulated robot holding the same object should exhibit similar proprioceptive responses (e.g., joint positions and velocities) when subjected to the same control inputs. 

The Algorithm \ref{alg:dhcem} describes the detail procedure of DH-CEM. In the initial iteration ($t = 0$), we generate $K$ hypotheses $\{ \mu_k^0 \}_{k=1}^K$ using estimates from a VLM and the object’s nominal size $\mathbf{s} = [a, b, c]^\top$. For each hypothesis, we draw $N/K$ parameter samples $\boldsymbol{\phi}_i^{(0)}$ by sampling the mass from a uniform distribution,
\[
m_i \sim \mathcal{U}((1 - \delta_m)\hat{m},\, (1 + \delta_m)\hat{m}),
\]
perturbing the nominal object size $\mathbf{s} = [a, b, c]^\top$ as
\[
\mathbf{s}_i \sim \mathcal{U}((1 - \delta_s)\mathbf{s},\, (1 + \delta_s)\mathbf{s}),
\]
and sampling the center of mass within the scaled object volume,
\[
\mathbf{h}_i \sim \mathcal{U}(-\delta_h \mathbf{s}_i,\, \delta_h \mathbf{s}_i).
\]
The inertia tensor \( \mathbf{I}_i \) is subsequently computed from the sampled mass, size, and center of mass, under the assumption of a uniform-density cuboid (see Eq. (\ref{eq:diag_inertia_tensor}) and Eq. (\ref{eq:offdiag_inertia_tensor})). The full tensor is constructed accordingly, with the parallel axis theorem applied when necessary.

Each $\boldsymbol{\phi}_i$ is injected into a physics simulator, and the system is rolled out using the same desired joint trajectory $\{ \mathbf{q}^*(t), \dot{\mathbf{q}}^*(t) \}_{t=1}^T$. The resulting simulated trajectory $\{ \hat{\mathbf{q}}_i(t), \hat{\dot{\mathbf{q}}}_i(t) \}_{t=1}^T$ is evaluated using the trajectory prediction error:
\begin{equation}
J(\boldsymbol{\phi}_i) = \sum_{t=1}^T \left\| \hat{\mathbf{q}}_i(t) - \mathbf{q}^*(t) \right\|^2 + \left\| \hat{\dot{\mathbf{q}}}_i(t) - \dot{\mathbf{q}}^*(t) \right\|^2,
\end{equation}
where \( \hat{\mathbf{q}}_i(t) \) and \( \hat{\dot{\mathbf{q}}}_i(t) \) denote the simulated joint position and velocity for the \( i \)-th hypothesis, and \( \mathbf{q}^*(t), \dot{\mathbf{q}}^*(t) \) are the reference trajectories.

\noindent The top \( \rho N \) hypotheses with the lowest cost define the elite set \( \mathcal{E}^{(0)} \), from which we compute the updated Gaussian parameters for mass and center of mass (CoM) as
\begin{equation}
\mu^{(1)} = \frac{1}{|\mathcal{E}^{(0)}|} \sum_{\boldsymbol{\phi} \in \mathcal{E}^{(0)}} \boldsymbol{\phi}, \quad
\sigma^{(1)} = \sqrt{ \frac{1}{|\mathcal{E}^{(0)}|} \sum_{\boldsymbol{\phi} \in \mathcal{E}^{(0)}} (\boldsymbol{\phi} - \mu^{(1)})^2 }.
\end{equation}

\noindent In subsequent iterations (\( t > 0 \)), new samples are generated from the updated Gaussian distribution \( \mathcal{N}(\mu^{(t)}, \sigma^{(t)}) \), while the object size remains fixed. The inertia tensor is then recomputed deterministically from the sampled mass and CoM. After a few iterations (three in our case), we obtain the estimated inertial parameters of the object, denoted by
\[
\widehat{\boldsymbol{\Phi}}_{\text{obj}} =
\begin{bmatrix}
\hat{m}_{\mathrm{obj}},\ 
\hat{\mathbf{h}}_{\mathrm{obj}}^\top,\ 
\hat{I}_{xx}^{\mathrm{obj}},\ 
\hat{I}_{yy}^{\mathrm{obj}},\ 
\hat{I}_{zz}^{\mathrm{obj}},\ 
\hat{I}_{xy}^{\mathrm{obj}},\ 
\hat{I}_{yz}^{\mathrm{obj}},\ 
\hat{I}_{zx}^{\mathrm{obj}}
\end{bmatrix}^\top.
\]




\begin{figure}[t]
    \centering
    \includegraphics[width=0.52\linewidth]{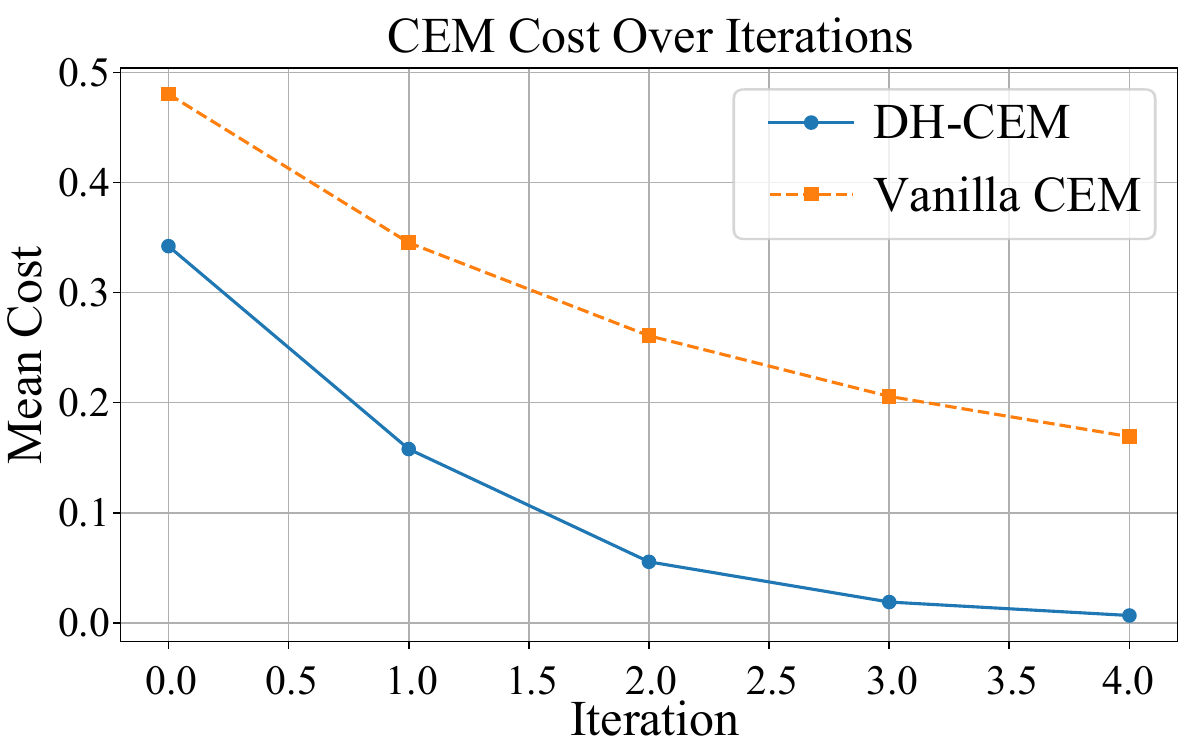}
    \includegraphics[width=0.45\linewidth]{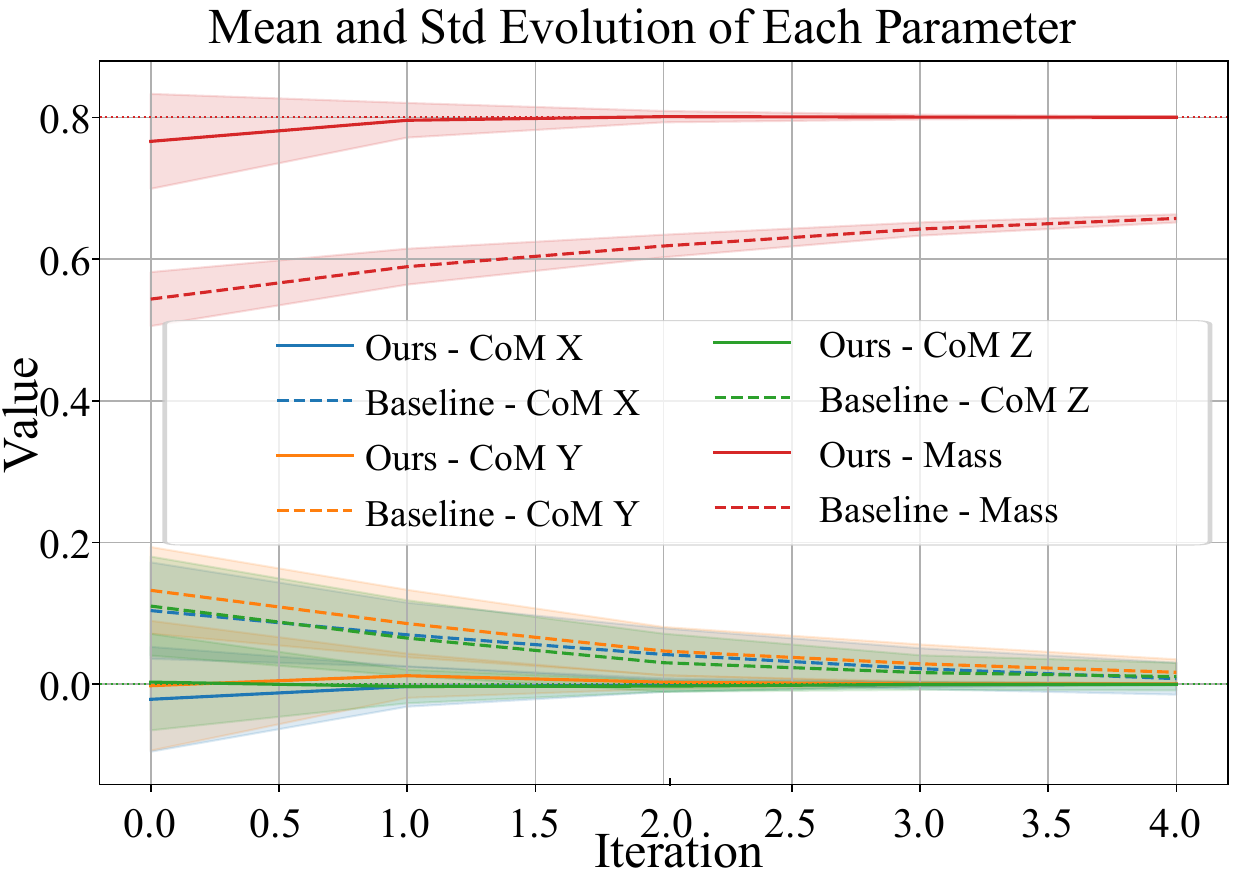}
    \caption{Change in cost and parameter convergence. (Top) Cost comparison between the baseline (vanilla CEM) and DH-CEM. (Bottom) Parameter distribution at convergence.}
    \label{fig:cem_results}
\end{figure}

\section{Object-Aware Whole-Body Bilateral Teleoperation}

The concept of dynamic similarity, which models both the human and robot as linear inverted pendulums and synchronizes their dynamics via haptic feedback and natural frequency, was introduced by Ramos et al.~\cite{ramos2019dynamic, ramos2018humanoid} and extended to wheeled humanoids in~\cite{purushottam2022hands, purushottam2023dynamic}. We build on this framework by incorporating linearized cart-pole error dynamics to account for height and pitch variations, together with a multi-stage object parameter estimation.

\rev{
Without explicitly estimating human intent, predictive control is not directly applicable in our setting. 
We therefore adopt a reactive control strategy that updates the system equilibrium online, which is well-suited for bilateral haptic teleoperation requiring low latency and immediate responsiveness. While predictive controllers (e.g., MPC) can improve anticipatory motion and trajectory optimization, they require intent estimation and introduce additional computational latency. In contrast, our reactive formulation prioritizes immediate human input and minimizes latency, which is critical for stable and transparent haptic interaction. The estimated object inertial parameters are used to update the equilibrium point, enabling adaptive posture adjustment during lifting and compensating for object-induced torques during manipulation.
}

\rev{In this section, we present (i) a dynamic locomotion retargeting strategy accounting for height and equilibrium variations, (ii) an arm mapping and object-aware compensation scheme, and (iii) a safety-critical manipulation controller based on control barrier functions with estimated object inertial parameters.}

\subsection{Dynamic Locomotion Retargeting Considering Height and Equilibrium Change}\label{subsec:DynSim}

\subsubsection{New Equilibrium Point Estimation} \label{method:new_des_pitch}
\begin{figure}[t]
\centering
\includegraphics[width=\linewidth]{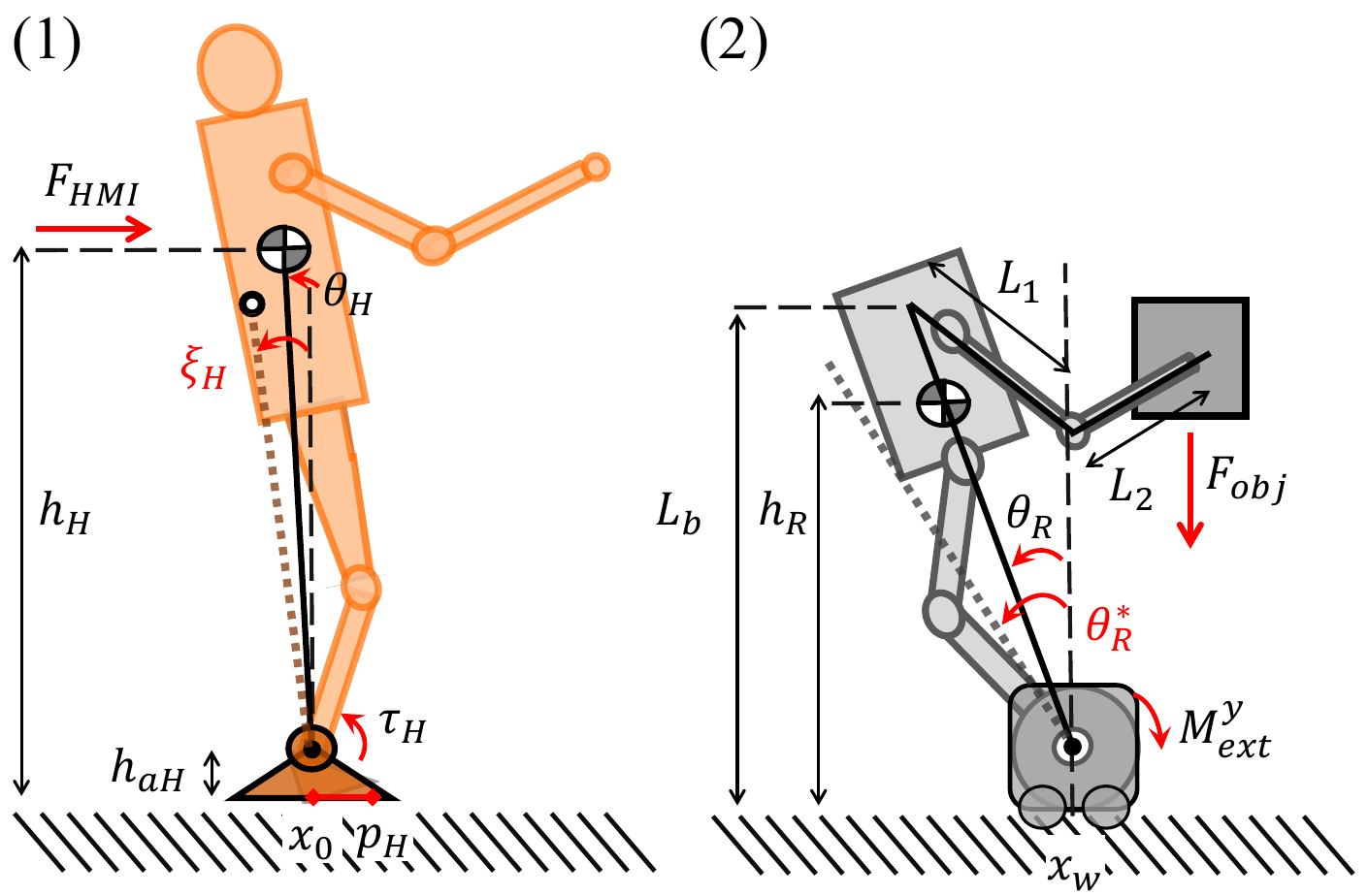}
\caption{Illustration of human and robot with reduced-order-model. The human is modeled as an actuated pendulum with ankle torque controlling lean, while the robot is represented as a cart-pole system for generating desired references.
}
\label{fig5_reduced_model}
\end{figure}

Lifting a heavy object induces a disturbance torque on the robot, shifting the system’s equilibrium point due to the added object dynamics. To maintain balance, the robot must adjust its pitch angle to a new static equilibrium that accounts for the shifted center of mass. We compute the corresponding desired pitch and pitch rate to enable stable tracking under the modified mass distribution (see Fig. \ref{fig5_reduced_model}-2). The moment balance around the wheel axis is:
\begin{equation} \label{eq:sum_of_moments}
    F_g h_R \sin(\theta_R^*) = -F_{\mathrm{obj}}\, x^x_{ee},
\end{equation}
where \( F_g \) and \( F_{\mathrm{obj}} \) denote the gravitational forces of the robot and the object, respectively, and \( x_{ee} \) is the horizontal position of the end-effector relative to the wheel axis (\( M^y_g\ = M^y_{ext}\)). We compute \( F_{\mathrm{obj}} \) using the estimated object mass \( \hat{m}_{\mathrm{obj}} \) as \( F_{\mathrm{obj}} = \hat{m}_{\mathrm{obj}} g \), where \( g \) is gravitational acceleration. The end-effector position is given by \( \mathbf{x}_{ee} = f(\mathbf{q}, \theta_R^*) \), where \( f \) maps the joint angles \(\mathbf{q}\) and base orientation \(\theta_R^*\) to \(\mathbf{x}_{ee}\). Substituting \( f(\mathbf{q}, \theta_R^*) \) into Eq.~\eqref{eq:sum_of_moments}, we solve for the equilibrium pitch angle \( \theta_R^* \) using:

\begin{equation} \label{eq:equil_pitch}
    \theta_R^* = \tan^{-1}\left(\frac{f_1}{f_2}\right),
\end{equation}
where
\begin{equation} \label{eq:f1_f2_def}
    f_1 = -F_{\mathrm{obj}} x^z_{ee}, \quad
    f_2 = F_g h_R + F_{\mathrm{obj}} x^x_{ee}
\end{equation}

\noindent To improve tracking performance, we compute the time derivative of Eq.~\eqref{eq:equil_pitch} to obtain the desired pitch rate:

\begin{equation} \label{eq:equil_pitch_vel}
    \dot{\theta}_R^* = \frac{f_2 \dot{f}_1 - f_1 \dot{f}_2}{f_1^2 + f_2^2},
\end{equation}

\noindent where \( \dot{f}_1 \) and \( \dot{f}_2 \) are functions of joint angles \( q_1, q_2 \), their velocities \( \dot{q}_1, \dot{q}_2 \), and the estimated object force \( F_{\mathrm{obj}} = \hat{m}_{\mathrm{obj}} g \). This formulation accounts for how dynamic arm motion alters the required lean angle for maintaining static equilibrium. Both \( \theta_R^* \) and \( \dot{\theta}_R^* \) are tracked by the height-varying LQR controller in Eq.~\eqref{eq:LQR_control}, whose gain is adaptively adjusted based on the current body height. \\

\subsubsection{Locomotion Retargeting of Error Dynamics}
Unlike prior work~\cite{purushottam2023dynamic}, which enforces direct DCM tracking via \( \xi_R = \xi_H \), we re-linearize the robot dynamics around a new equilibrium that accounts for object dynamics. This allows the robot adaptively regulating its lean posture by tracking a redefined desired DCM aligned with the shifted equilibrium. Tracking the human's DCM enables more dynamic robot behavior by capturing non-minimum phase characteristics as feedforward terms, while explicitly modeling external forces acting on both the human and robot. The desired pitch angle is:
\begin{equation}
    \xi_{R}^* = \theta_{R}^* + \frac{\dot{\theta}_{R}^*}{\omega_R}.
\end{equation}

\noindent To minimize the DCM tracking error while accounting for size and dynamics mismatch between the human and robot, we constrain the normalized DCM derivative difference~\cite{ramos2018humanoid}:
\begin{equation}
\frac{\Delta \dot{\zeta}}{\omega_R} = \frac{\dot{\xi}_H}{\omega_H},
\quad \text{where } \Delta \dot{\zeta} := \dot{\xi}_R - \dot{\xi}_R^*.
\label{eq:dcm_delta_constraint}
\end{equation}

\noindent This constraint synchronizes the rate of DCM evolution across the human and robot. The DCM derivatives are defined as:
\[
\dot{\xi}_R = \dot{\theta}_R + \frac{\ddot{\theta}_R}{\omega_R}, \quad
\dot{\xi}_H = \dot{\theta}_H + \frac{\ddot{\theta}_H}{\omega_H}.
\]

We model the human as a fixed-base actuated inverted pendulum, where the ankle applies torque to regulate torso pitch, and the external force \( F_{\mathrm{HMI}} \) is applied at the CoM via a human–machine interface. The robot is modeled as a cart-pole system, where the base applies horizontal force \( F_R \) to control the pendulum's pitch, and an external disturbance \( F_{\mathrm{ext}} \) may act on the CoM. Both models assume small-angle linearization (\( \sin(\theta) \approx \theta \)) and uniform mass distribution, allowing us to treat each as a simple pendulum. Under these assumptions, the linearized pendulum dynamics for the robot and human are given by:
\begin{equation}
\ddot{\theta}_R = \omega_R^2 \theta_R - \frac{F_R}{\gamma_R} + \frac{F_{\mathrm{ext}}}{\alpha^2 \gamma_R}
\end{equation}
\begin{equation}
\ddot{\theta}_H = \omega_H^2 \theta_H - \frac{p_H}{h_H} + \frac{F_{\mathrm{HMI}}}{\gamma_H},
\end{equation}
where \( \omega_j = \sqrt{g / h_j} \) is the natural frequency of the pendulum and \( \gamma_j = m_j \omega_j^2 h_j \) for \( j \in \{H, R\} \) is a nondimensionalizing gain with units of force. The term \( \alpha^2 = M / m \) reflects the relative mass ratio between the robot's base and pendulum. Substituting these expressions into the DCM derivative definitions,
\[
\dot{\xi}_R = \dot{\theta}_R + \frac{\ddot{\theta}_R}{\omega_R}, \quad 
\dot{\xi}_H = \dot{\theta}_H + \frac{\ddot{\theta}_H}{\omega_H},
\]
we obtain:
\begin{equation}
\dot{\xi}_R = \dot{\theta}_R + \omega_R \theta_R - \frac{F_R}{\omega_R \gamma_R} + \frac{F_{\mathrm{ext}}}{\omega_R \alpha^2 \gamma_R}
\end{equation}
\begin{equation}
\dot{\xi}_H = \dot{\theta}_H + \omega_H \theta_H - \frac{p_H}{\omega_H h_H} + \frac{F_{\mathrm{HMI}}}{\omega_H \gamma_H},
\end{equation}

\noindent These expressions show how each DCM evolves as a function of pitch angle, pitch velocity, and external inputs. Notably, the terms involving \( F_R \), \( F_{\mathrm{ext}} \), \( F_{\mathrm{HMI}} \), and \( p_H \) characterize how actuation and interaction forces shape the DCM dynamics for both human and robot. Plugging into Eq.~\eqref{eq:dcm_delta_constraint} and rearranging, we obtain the following constraint that couples the human and robot angular dynamics:
\begin{equation}
\theta_R + \frac{\dot{\theta}_R}{\omega_R} 
- \frac{F_R}{\gamma_R} + \frac{F_{\mathrm{ext}}}{\alpha^2 \gamma_R}
= \theta_H + \frac{\dot{\theta}_H}{\omega_H} 
- \frac{p_H}{h_H} + \frac{F_{\mathrm{HMI}}}{\gamma_H}
- \frac{\Delta \dot{\zeta}}{\omega_R},
\label{eq:dcm_constraint_deltazeta}
\end{equation}

\noindent where \( \gamma_j = m_j \omega_j^2 h_j \) for \( j \in \{H, R\} \) is a nondimensionalizing coefficient with units of force. This relationship enables coordinated DCM regulation and allows for explicit compensation of human interaction forces and environmental disturbances.

To enforce this constraint, we define the feedback force to the human and the feedforward control to the robot as:
\begin{align}
F_{\mathrm{HMI}} &= \gamma_H \left( (\theta_R - \theta_H) + \left( \frac{\dot{\theta}_R}{\omega_R} - \frac{\dot{\theta}_H}{\omega_H} \right) 
+ \frac{F_{\mathrm{ext}}}{\alpha^2 \gamma_R} + \frac{\Delta \dot{\zeta}}{\omega_R} \right) \\
F_R &= \gamma_R  \frac{p_H}{h_H},
\label{eq:feedback_feedforward}
\end{align}
where \( \gamma_j = m_j \omega_j^2 h_j \) for \( j \in \{H, R\} \) are nondimensionalizing gains, and \( \Delta \dot{\zeta} := \dot{\xi}_R - \dot{\xi}_R^* \) denotes the DCM rate mismatch. In practice, we assume the human's height changes slowly relative to the controller's update rate, allowing it to be treated as constant within each control iteration, with a fixed natural frequency. \\

\subsubsection{Locomotion Retargeting for Height Change}\label{subsec:DynSim}

To regulate balance during height transitions, we employ a feedback controller that combines gain-scheduled LQR with feedback linearization:
\begin{equation} \label{eq:LQR_control}
    u = -\boldsymbol{K}_{\mathrm{LQR}}(h_R)(\boldsymbol{q}_{xR}^{\mathrm{des}} - \boldsymbol{q}_{xR}) - \boldsymbol{B}^{\dagger} \boldsymbol{d}_w,
\end{equation}
where \( \boldsymbol{q}_{xR} \) is the robot’s reduced-order state, \( \boldsymbol{K}_{\mathrm{LQR}}(h_R) \) is linearly interpolated based on robot height, and \( \boldsymbol{B}^{\dagger} \) denotes the Moore–Penrose pseudo-inverse of the actuation matrix. The disturbance \( \boldsymbol{d}_w \in \mathbb{R}^2 \) compensates for base forces arising from leg motion:
\begin{equation}
    \boldsymbol{d}_w = 
    \begin{bmatrix}
        F_L \sin(\theta_R) \\
        0
    \end{bmatrix},
\end{equation}
where \( F_L \) is the estimated ground reaction force and \( \theta_R \) is the robot pitch. To enable height modulation, we map human height changes to desired robot height using:
\begin{equation}
    h_R^{\mathrm{des}} = h_R^{\mathrm{nom}} + \beta_z \frac{h_R^{\mathrm{nom}}}{h_H^{\mathrm{nom}}}(h_H - h_H^{\mathrm{nom}}),
\end{equation}
where \( \beta_z \in [0,1] \) tunes the mapping sensitivity. This approach provides intuitive vertical control without explicitly modeling $z$-axis dynamics. \\

\subsection{Manipulation Control and Arm Mapping Strategy} \label{subsec:Arm_Mapping}

\subsubsection{Kinematic Manipulation Retargeting}
We implement a kinematic retargeting method \cite{purushottam2023dynamic} that maps human arm motion, captured via a 4-DoF exosuit of HMI, to a 4-DoF SATYRR's manipulator. While the robot shoulder is modeled as a true spherical joint with intersecting axes, the exosuit's shoulder configuration consists of non-intersecting rotational axes. To resolve this mismatch, we use a projection-based approach to construct a feasible robot shoulder orientation.

Let \( q^H, q^R \in \mathbb{R}^4 \) denote the human and robot joint angles, respectively. The direction from the shoulder to elbow, \( \mathcal{R}^z_H \), is computed from the exosuit’s forward kinematics and is aligned with the robot’s elbow direction \( \mathcal{R}^z_R = \mathcal{R}^z_H \). To complete the shoulder orientation, we define the plane
\begin{equation}
    \mathcal{S} = \left\{ \mathcal{R}_u \;|\; \mathcal{R}_u^\top \mathcal{R}^z_H = 0 \right\},
\end{equation}
orthogonal to \( \mathcal{R}^z_H \) and project the human’s elbow rotation axis \( \mathcal{R}^y_H \) onto this plane to obtain the robot’s second shoulder axis:
\begin{equation}
    \mathcal{R}^y_R = \mathrm{proj}_{\mathcal{S}}(\mathcal{R}^y_H),
\end{equation}
followed by the cross product \( \mathcal{R}^x_R = \mathcal{R}^y_R \times \mathcal{R}^z_R \) to complete the orthonormal basis. The first three robot joint angles are then solved via inverse kinematics:
\begin{equation}
    q^R_{[0:2]} = \mathrm{IK}(\mathcal{R}^x_R, \mathcal{R}^y_R, \mathcal{R}^z_R),
\end{equation}
while the elbow angle is directly mapped: \( q^R_3 = q^H_3 \). This approach enables spatially consistent and intuitive retargeting from human to robot despite structural differences. \\



\subsubsection{Object-Aware Manipulation Control}
To enable accurate manipulation under varying object dynamics while maintaining compliant behavior for safe human–robot interaction, we employ a control strategy that combines a proportional-derivative (PD) feedback controller with a feedforward compensation term derived from inverse dynamics. Specifically, we use low PD gains to ensure compliance during physical interaction, and compensate for the object-induced dynamics using estimated inertial parameters \( \widehat{\boldsymbol{\Phi}}_{\text{obj}} \). The control input is defined as:
\begin{equation}
\boldsymbol{\tau} = \mathbf{K}_p (\mathbf{q}_d - \mathbf{q}) + \mathbf{K}_d (\dot{\mathbf{q}}_d - \dot{\mathbf{q}}) + \widehat{\mathbf{M}}(\mathbf{q}) \ddot{\mathbf{q}}_d + \widehat{\mathbf{b}}(\mathbf{q}, \dot{\mathbf{q}}),
\label{eq:pd_inverse_dynamics}
\end{equation}

\noindent
where \( \mathbf{K}_p \) and \( \mathbf{K}_d \) are low PD gain matrices to preserve compliance, and \( \widehat{\mathbf{M}}(\mathbf{q}) \) and \( \widehat{\mathbf{b}}(\mathbf{q}, \dot{\mathbf{q}}) \) denote the estimated inertia matrix and nonlinear effects (Coriolis, centrifugal, and gravity), respectively, based on \( \hat{\boldsymbol{\Phi}}_{\text{obj}} \). This formulation enables the robot to adapt its motion and torque commands according to the physical properties of the manipulated object, achieving both precise tracking and compliant behavior necessary for safe and effective human–robot interaction. \\

\subsubsection{Application of Object-Aware Robust Adaptive Safe Manipulation with Control Barrier Functions} Separate from the teleoperation setting, we also demonstrate our multi-stage object parameter estimator in an autonomous manipulation context, with a focus on collision avoidance. We incorporate a control barrier function (CBF)-based safety filter \cite{Ames2016} into a null-space task-space manipulation controller. To address modeling uncertainty in the object's dynamics which can lead to violation of the CBF constraint, we leverage our multi-stage sampling-based estimator to adapt the constraint against variations introduced by the manipulated object. Based on~\cite{khazoom2022humanoid}, we formulate the CBF constraint at the acceleration level to integrate the full-body dynamics of both the robot and the object within a quadratic program (QP) framework:

\begin{equation}
\begin{aligned}
\min_{\boldsymbol{\tau}, s_{\text{cbf}}} \quad & \frac{1}{2} \boldsymbol{\tau}^\top H \boldsymbol{\tau} + \mathbf{f}^\top \boldsymbol{\tau} + \lambda_{\text{cbf}} \|s_{\text{cbf}}\|_1 \\
\text{s.t.} \quad 
& \ddot{\mathbf{q}} = f_{\mathrm{FD}}(\mathbf{q}, \dot{\mathbf{q}}, \boldsymbol{\tau}; \widehat{\boldsymbol{\Phi}}_{\text{robot}}, \widehat{\boldsymbol{\Phi}}_{\text{obj}}), \\
& \boldsymbol{\tau}_{\min} \leq \boldsymbol{\tau} \leq \boldsymbol{\tau}_{\max}, \\
& \text{CBF constraint for collision avoidance~\eqref{eq:cbf_constraint}}
\end{aligned}
\label{eq:object_aware_qp}
\end{equation}

\noindent where the objective minimizes a weighted deviation of the torque command $\boldsymbol{\tau}$ via a quadratic cost defined by $H$ and $\mathbf{f}$, while penalizing the slack variable $s_{\text{cbf}}$ through $\ell_1$-norm terms weighted by $\lambda_{\text{cbf}}$. The slack variable allows minimal relaxation of the safety constraint when necessary. The equality constraint enforces the robot’s forward dynamics given the current joint state $(\mathbf{q}, \dot{\mathbf{q}})$ and the estimated inertial parameters of both the robot $\widehat{\boldsymbol{\Phi}}_{\text{robot}}$ and the object $\widehat{\boldsymbol{\Phi}}_{\text{obj}}$. Torque saturation limits $\boldsymbol{\tau}_{\min}$ and $\boldsymbol{\tau}_{\max}$ are imposed as hard constraints. Finally, collision avoidance is enforced via a robust Control Barrier Function (CBF) constraint at the acceleration level, softened by the slack variable $s_{\text{cbf}}$:
\begin{equation}
\begin{aligned}
&\mathbf{J}_{AB} \ddot{\mathbf{q}} 
+ \dot{\mathbf{J}}_{AB} \dot{\mathbf{q}} 
+ \alpha_1^{AB} \mathbf{J}_{AB} \dot{\mathbf{q}} \\
&\quad + \alpha_1^{AB} \alpha_2^{AB} h_{AB}^{\text{sd}} 
\geq \epsilon(h_{AB}^{\text{sd}}) - s_{\text{cbf}}, 
\quad \forall AB \in \mathcal{P},
\end{aligned}
\label{eq:cbf_constraint}
\end{equation}

\noindent where $\mathbf{J}_{AB}$ is the relative Jacobian between links $A$ and $B$, $h_{AB}^{\text{sd}}$ is the signed distance function, and $\alpha_1^{AB}, \alpha_2^{AB} \in \mathbb{R}_+$ are class-$\mathcal{K}$ parameters shaping responsiveness. Here, $h_{AB}^{\text{sd}}$ serves as a CBF \cite{nguyen2016exponential} with relative-degree two. The robust margin $\epsilon$ is designed as
\begin{equation}
\epsilon(h_{AB}^{\text{sd}}) := \epsilon_1 +\frac{k}{\sqrt{(h_{AB}^{\text{sd}})^2 + \epsilon_2}},
\label{eq:robust_margin}
\end{equation}

\noindent based on the Input-to-state safe robust CBF constraint formulation \cite{alan2023control}, which mitigates the discrepancy between the estimated and the actual dynamics of the robot. The set $\mathcal{P}$ comprises all body pairs for which safety must be enforced. 

In summary, the adaptation of the CBF constraint is achieved in Eq. \eqref{eq:object_aware_qp} through using the estimated parameters in $f_{\mathrm{FD}}$, and the robustness is achieved through the term $\epsilon$ in Eq. \eqref{eq:cbf_constraint}, which provides an appropriate level of conservatism to address uncertainty while ensuring feasibility in real time.

%% file: table/centersnap_table.tex
\begin{table}[t]
\centering
\caption{Estimated vs. true object dimensions (in mm) for various bottle-like objects. While height is generally estimated accurately, width and length are often underestimated for larger objects. This systematic bias is treated as uncertainty and considered in the third stage of the multi-stage estimation process, which employs a sampling-based method.}

\resizebox{0.98\linewidth}{!}{%
\begin{tabular}{lccc}
\toprule
\textbf{Object} & \textbf{Width} & \textbf{Length} & \textbf{Height} \\
\midrule
Green Water Bottle & 72.2 / 80.3 & 72.2 / 85.0 & 245.0 / 258.0 \\
Stock Water Bottle & 97.1 / 81.0 & 97.2 / 82.0 & 250.0 / 267.0 \\
White Thermal Cup & 89.2 / 85.0 & 89.2 / 81.0 & 185.0 / 215.0 \\
Cylindrical Water Bottle & 73.6 / 81.0 & 73.6 / 82.0 & 240.1 / 271.0 \\
Plastic Water Bottle & 70.5 / 74.1 & 70.5 / 76.0 & 233.1 / 257.0 \\
Plastic Big Water Bottle & 145.0 / 100.1 & 145.0 / 93.4 & 240.0 / 240.0 \\
Big Water Bottle + Box & 150.0 / 110.0 & 150.0 / 110.0 & 250.0 / 250.0 \\
I shape object & 78.2 / 100.0 & 78.3 / 50.0 & 206.9 / 190.0 \\
\bottomrule
\end{tabular}
}
\label{tab:size_estimation_compact}
\end{table}

%% file: Experiment.tex
\section{Results}
\label{Results}

The proposed whole-body bilateral teleoperation framework, integrated with a multi-stage, sampling-based online object inertial parameter estimation method, is evaluated on the wheeled humanoid SATYRR controlled by our customized HMI. A total of five experiments are conducted in both simulation and hardware. Details of the hardware setup are described in Appendix~\ref{appendix::system} (see Fig. \ref{fig3_system_description}).

First, in Section~\ref{VI-A}, we benchmark the proposed multi-stage object inertial parameter estimation method using a pre-recorded offline dataset consisting of diverse arm trajectories and manipulated objects (see Fig.~\ref{fig3_cetersnap}). This quantitatively evaluates estimation accuracy and isolates the contributions of key components, including vision and vision-language model (VLM) priors, a decoupled hierarchical structure, and sim-to-real adaptation. \rev{We further evaluate estimation accuracy across objects with varying degrees of deviation from cuboid geometry to analyze how the cuboid assumption holds or degrades.}

Second, in Section~\ref{VI-A3}, we assess the estimator’s real-time adaptability in simulation by dynamically varying the object’s ground-truth inertial parameters. This includes sim-to-sim validation, where we generate sample trajectories in Isaac Gym and test the estimator on a target robot model in MuJoCo, demonstrating robustness to cross-simulator discrepancies and responsiveness to online changes in object properties.

Third, we deploy the proposed framework in hardware experiments involving standard pick-and-place tasks (Section~\ref{VI-B1}) and the delivery of heavy objects through lifting, squatting, and releasing motions (Section~\ref{VI-B2}). These experiments evaluate whole-body loco-manipulation performance and the role of haptic force feedback in human-in-the-loop control. Accurate object parameter estimation is critical in this context, as estimation errors are reflected in the haptic feedback, degrading the operator’s ability to compensate for object dynamics.

Fourth, in Section~\ref{VI-B3}, we \rev{integrate the estimated parameters into a model-based safety-critical controller using control barrier functions (CBFs) for collision avoidance, demonstrating a downstream application of the proposed estimator beyond teleoperation.}

Finally, we analyze the impact of sim-to-real adaptation
across varying action spaces and parameter accuracies, underscoring the role of high-fidelity simulation in enhancing sampling-based estimation effectiveness. 
\rev{In addition, we conduct controlled ablation studies to evaluate robustness to residual model mismatch and sensitivity to trajectory excitation. Further details are provided in Appendix~\ref{appendix::sim-to-real}--\ref{appendix::excitation}.}

\subsection{Evaluation of Object Inertial Parameter Estimation}

\begin{table*}[t]
\centering
\small
\caption{Estimation results of the vision-language model (VLM) and our multi-stage object parameter estimation framework. The table compares outputs from the VLM (which includes Stage 1 and 2) with our complete method (including Stage 3), highlighting incremental performance improvements throughout physical interaction. Stage 3 is based on the DH-CEM approach, which refines the initial estimates. Values of the CoM are reported in millimeters, and inertia is represented as vectors scaled by $10^3$. Improvements in inertia are evaluated using the L2 norm.} 
\label{tab:percentage_improvement}
\begin{tabular}{l|ccc|c|cc|c|ccc|c}
\toprule
\textbf{Object} & \multicolumn{3}{c|}{\textbf{Mass (kg)}} & \textbf{Mass} & \multicolumn{2}{c|}{\textbf{CoM (mm)}} & \textbf{CoM} & \multicolumn{3}{c|}{\textbf{Inertia ($\times10^3$ kg·m$^2$)}} & \textbf{Inertia} \\
 & GT & VLM & Ours & \textbf{Imp. (\%)} & VLM & Ours & \textbf{Imp. (\%)} & GT & VLM & Ours & \textbf{Imp. (\%)} \\
\midrule
coke            & 0.655 & 1.160 & 0.791 & 73.1 & 8.0 & 5.1 & 36.3 & 3,3,1 & 7,7,1 & 5,5,1 & 50.0 \\
cylindrical     & 0.830 & 1.440 & 0.873 & 93.0 & 5.4 & 3.0 & 44.4 & 4,4,1 & 8,8,1 & 5,5,1 & 75.0 \\
stock\_water    & 1.580 & 1.420 & 1.581 & 99.4 & 8.5 & 12.2 & -43.5 & 10,10,2 & 10,10,2 & 11,10,2 & - \\
full\_steel     & 0.743 & 0.670 & 0.752 & 87.7 & 1.4 & 0.0 & 100.0 & 3,1,3 & 1,1,0 & 1,1,1 & 21.6 \\
half\_and\_half & 0.523 & 0.670 & 0.662 & 5.4 & 12.1 & 11.1 & 8.3 & 2,1,2 & 3,3,0 & 3,3,0 & 0.0 \\
barbell         & 0.567 & 0.670 & 0.514 & 48.5 & 2.2 & 2.2 & 0.0 & 3,1,3 & 1,1,1 & 2,1,1 & 20.9 \\
corner          & 0.464 & 0.670 & 0.397 & 67.5 & 4.1 & 7.0 & -70.7 & 2,1,2 & 3,4,1 & 2,3,0 & 14.7 \\
empty           & 0.228 & 0.670 & 0.428 & 54.8 & 4.2 & 7.8 & -85.7 & 1,0,1 & 2,3,2 & 2,2,1 & 32.6 \\
half\_abs       & 0.266 & 0.670 & 0.412 & 63.9 & 2.0 & 1.4 & 30.0 & 1,0,1 & 1,1,0 & 1,1,0 & 0.0 \\
\bottomrule
\end{tabular}
\end{table*}

\subsubsection{Benchmarking Parameter Estimation on Hardware}
\label{VI-A}
\rev{We benchmark our object parameter estimation using real-world data, comparing it against ablated variants of our method, a physically feasible optimization-based approach \cite{rucker2022smooth}, and representative methods from different classes, including optimization-based, regression-based, and learning-based approaches.}

The ground truth dataset consists of 30 samples, covering 10 objects with 3 trajectories each (see Fig.~\ref{fig3_cetersnap}), collected using the SATYRR platform and a human–machine interface (HMI). A human pilot teleoperated the arm using natural, user-preferred motions without predefined or excitation-optimized trajectories \cite{baek2024online}.

We evaluate the following baselines:
\begin{itemize}
    \item \textbf{\textit{NV-NVLM}}: No visual or VLM priors.
    \item \textbf{\textit{WV-NVLM}}: Visual size only.
    \item \textbf{\textit{WV-WVLM}}: Visual size + VLM priors.
    \item \textbf{\textit{Chol. LS}}: Physically constrained least-squares-based method \cite{wensing2017linear, rucker2022smooth}.
    \item \rev{\textbf{\textit{Grid Sweep}}}: Grid search over $(m, c_x, c_y, c_z)$.
    \item \rev{\textbf{\textit{Bayes. Opt.}}}: Bayesian optimization under identical initialization.
    \item \rev{\textbf{\textit{L-BFGS-B}}}: Gradient-based optimization on trajectory matching.
    \item \textbf{\textit{Ours}}: Full DH-CEM method.
    \item \textbf{\textit{Ours w/o SysID}}: Without sim-to-real adaptation.
    \item \rev{\textbf{\textit{Learning-based}}: A learning-based inertial parameter estimator trained on simulated interaction data \cite{baek2024online}, where the training distribution is constructed to cover the range of object sizes and masses observed in our evaluation set.}
\end{itemize}

\input{table/obj_shape_table}

\begin{figure*}[t]
\centering
 \includegraphics[width=\textwidth]{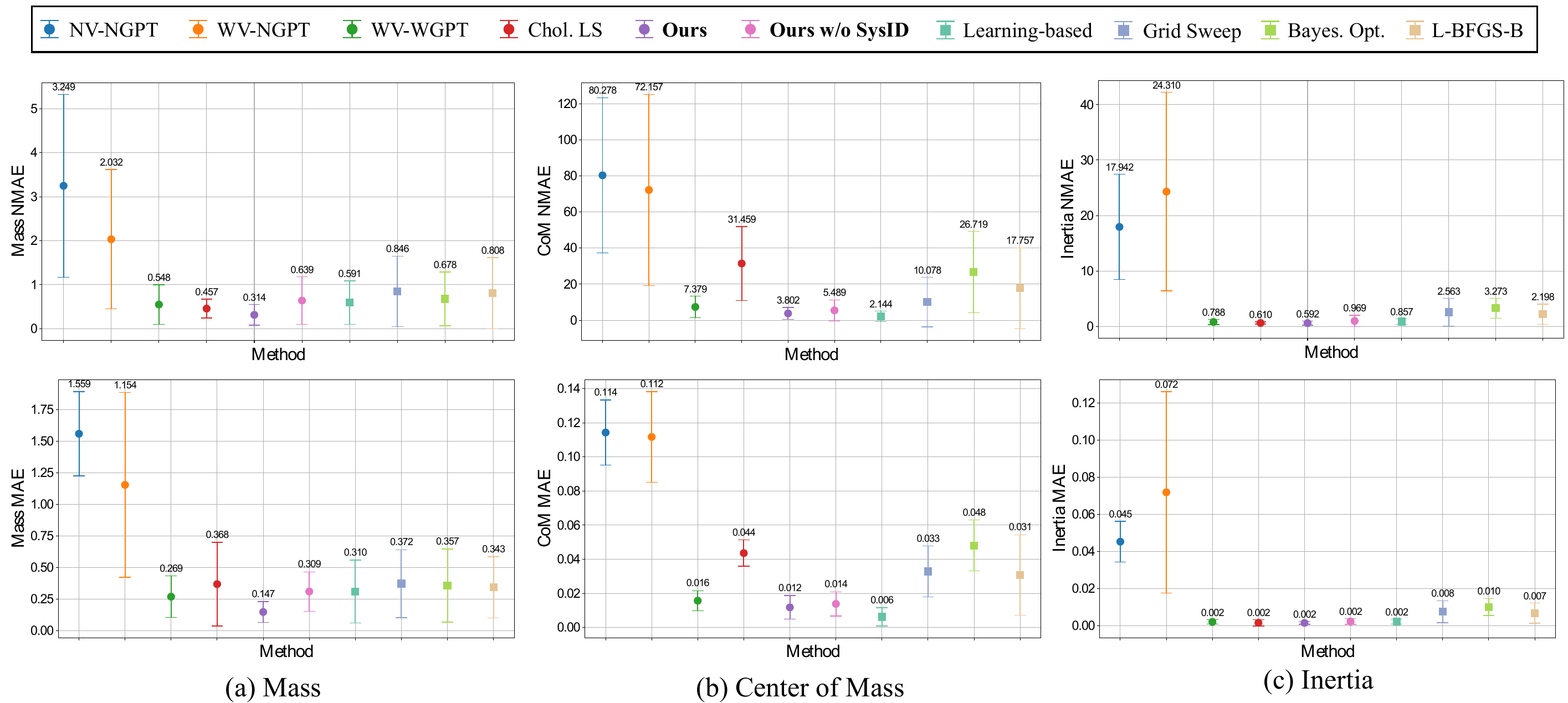}
 \vspace{-1.5em}
    \caption{\rev{
    Benchmark results of object inertial parameter estimation on real-world data. 
    Each bar shows the mean and standard deviation of normalized (top) and unnormalized (bottom) mean absolute errors (MAE) across mass, center of mass, and inertia. 
    The comparison includes ablated variants, optimization-based baselines, a learning-based estimator, and alternative optimizers (Grid Sweep, Bayesian Optimization, L-BFGS-B) under identical VLM initialization. 
    \revtwo{DH-CEM with all priors shows the best overall performance. The vision-language priors and geometric decomposition reduce the number of parameters that must be identified from interaction data, while the decoupled-hierarchical CEM improves convergence speed and search efficiency within that reduced space.}
    }}
    \label{fig9}
    \vspace{-0.5em}
\end{figure*}

As shown in Fig.~\ref{fig9}, DH-CEM with all priors (Ours) achieves consistently low Normalized Mean Absolute Error (NMAE) and Mean Absolute Error (MAE) across mass, center of mass (CoM), and inertia, demonstrating strong overall performance across different parameter types. \rev{While the learning-based estimator achieves competitive performance in CoM estimation, it exhibits higher variance and reduced accuracy in inertia estimation compared to the proposed method.}

\begin{figure}[t]
\centering
\includegraphics[width=1\linewidth]{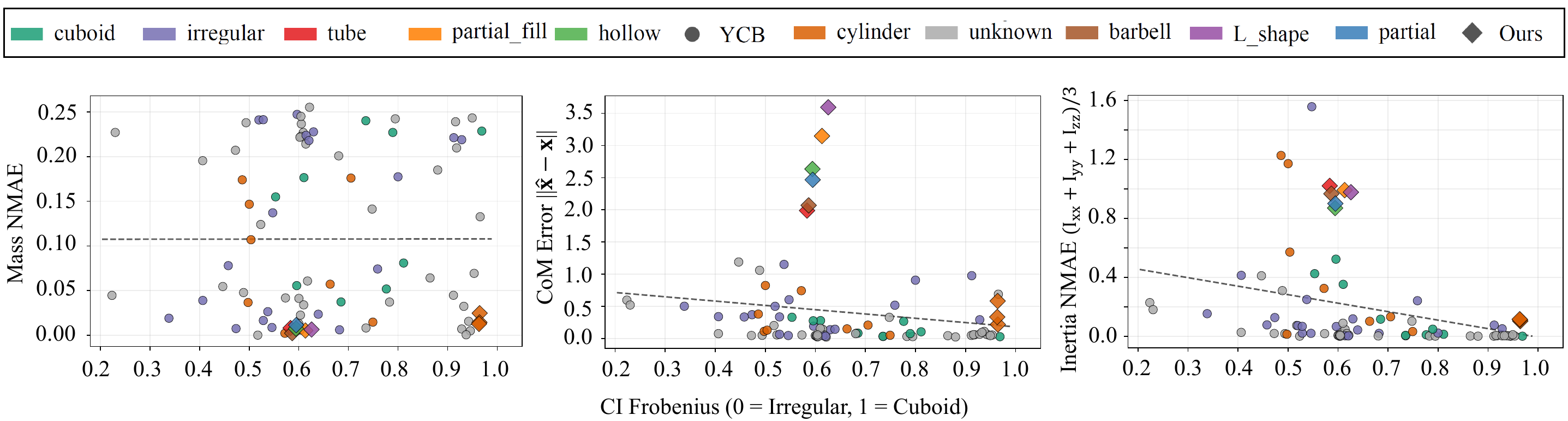} 
\caption{\rev{Estimation performance versus cuboidness index (CI). Each point represents an object instance. Errors for mass (left), CoM (middle), and inertia (right) are shown. Performance degrades as objects deviate from cuboid geometry (lower CI), with CoM and inertia more affected than mass.}}
\label{fig:abl_shape}
\end{figure}

Incorporating prior knowledge from vision (object size) and vision-language models (VLM) substantially improves estimation performance by providing informative initial guesses of inertial parameters. 
The decoupled hierarchical structure contributes to more stable and efficient convergence by sequentially estimating mass, CoM, and inertia, thereby reducing parameter coupling. In addition, the use of multivariate initialization increases robustness against poor priors by exploring multiple hypotheses in parallel. High-fidelity simulation further enhances performance by enabling accurate and consistent evaluation of sampled parameters. \revtwo{Under identical VLM-based initialization, DH-CEM achieves consistently lower error while converging in approximately 0.3--0.5~s, compared to $\sim$4~s (Grid Sweep), $\sim$13~s (Bayesian Optimization), and $\sim$3~s (L-BFGS-B). This gap reflects differences in search efficiency over the same information, rather than differences in excitation requirements} 

\rev{
Table~\ref{tab:percentage_improvement} summarizes the estimated parameters across different objects. 
Mass estimation improves substantially for most objects (e.g., \textit{cylindrical}: 93.0\%, \textit{full steel}: 87.7\%), while objects with eccentric mass distributions such as \textit{barbell} show limited gain, as the VLM prior already provides a reasonable estimate. In contrast, CoM estimation exhibits more variable behavior and can degrade after Stage~3 for certain objects (e.g., \textit{stock water}: $-43.5\%$, \textit{corner}: $-70.7\%$, \textit{empty}: $-85.7\%$), indicating sensitivity to object geometry and potential model mismatch. This suggests that the effectiveness of interaction-based refinement depends on the validity of the underlying geometric assumption.}

\rev{
\subsubsection{Sensitivity to Cuboid Assumption}
\label{VI-A2}
To analyze sesitivity to violations of the cuboid assumption, we conduct additional simulation experiments across objects with diverse geometries, including both custom-designed objects and YCB dataset instances. Geometric deviation is quantified using a cuboidness index (CI), defined as a normalized Frobenius-based similarity to a best-fit cuboid (1: perfect cuboid, 0: highly irregular). As shown in Table~\ref{tab:cuboidness_shape} and Fig.~\ref{fig:abl_shape}, estimation accuracy degrades with lower CI, particularly for CoM and inertia, while mass remains relatively stable. In particular, objects with irregular geometry or non-uniform internal mass distribution exhibit larger CoM and inertia errors, whereas cuboid and near-cuboid objects remain accurate. Overall, the method is most reliable for CI $> 0.65$, with performance degrading for highly non-cuboidal objects (e.g., barbells, L-shaped), highlighting a key limitation.
}

\begin{figure}[t]
\centering
\includegraphics[width=0.95\linewidth]{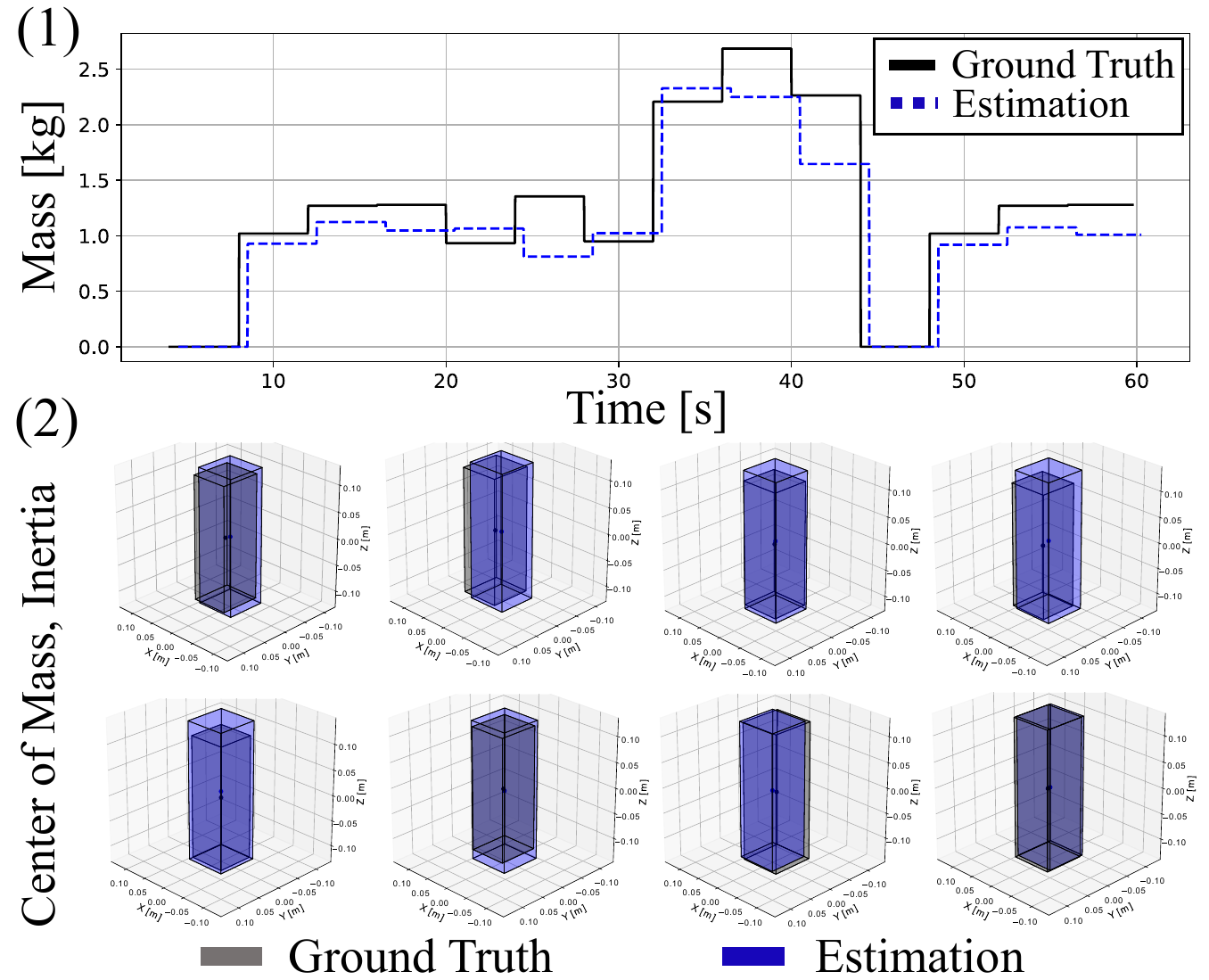}
\caption{Real-time online parameter estimation performance in simulation. (1) Mass tracking results comparing the ground-truth mass (black) and the estimated mass (blue dashed), illustrating the estimator's ability to follow time-varying object properties. (2) Comparison between the estimated and ground-truth object center of mass (CoM) and inertia, visualized using the object's geometry at selected time steps, demonstrating accurate adaptation to changing physical parameters.
}
\label{real_time_param_est}
\end{figure}

\subsubsection{Real-Time Online Parameter Estimation in Simulation}
\label{VI-A3}
We validate the online performance of our multi-stage object inertial parameter estimation in simulation by dynamically varying the ground-truth object parameters in real time, including sim-to-sim validation using data generated in Isaac Gym and tested on a target robot in MuJoCo. Each simulation environment operates in its own thread, communicating over local UDP. To leverage the estimated object size and VLM output, we apply the same object properties used in the hardware experiments. \rev{A predefined down-to-up trajectory, representative of natural lifting motions 
rather than dedicated excitation design, is used when the object changes.} For instance, when the ground truth inertial parameters of the manipulated object vary at 4-second intervals, the robot triggers the object parameter estimation process through a detection function activated by its arm-shaking motion.

The experimental results are shown in Fig.~\ref{real_time_param_est}. The estimation process takes approximately 0.5 seconds, with parameters remaining constant until the object changes. For clearer interpretation of estimation performance, we visualize the estimated cuboid and its center of mass. While the estimation is not perfectly accurate, it adequately supports human teleoperation by handling low-level control to adapt to dynamic changes, and enables collision avoidance through a model-based control barrier function during autonomous operation. \\

\subsection{Evaluation of Manipulation and Loco-Manipulation Tasks}

\subsubsection{Object Lifting Manipulation on Hardware}
\label{VI-B1}
To evaluate the impact of incorporating object physical parameter estimation in control, we conducted a heavy water bottle pick-and-place task. The objective was to lift a relatively heavy object and place it at an elevated location while maintaining compliant behavior to support safe and effective human-robot collaboration. We evaluated three scenarios: (1) comparing pick-and-place performance with and without object dynamics compensation; (2) manually moving the manipulator to assess its compliant behavior (see attached video); and (3) performing repeated pick-and-place tasks with varying target heights to test various scenarios (see attached video). To compensate for the unknown object's dynamics, its inertial parameters were estimated during the lifting phase and used to compute feedforward torques via inverse dynamics (Eq. \eqref{eq:pd_inverse_dynamics}). Compliant behavior was achieved by applying low PD gains at each joint.

\begin{figure*}[t]
    \centering

    \subfloat[\label{fig:box_img}]{
        \includegraphics[width=0.6\textwidth]{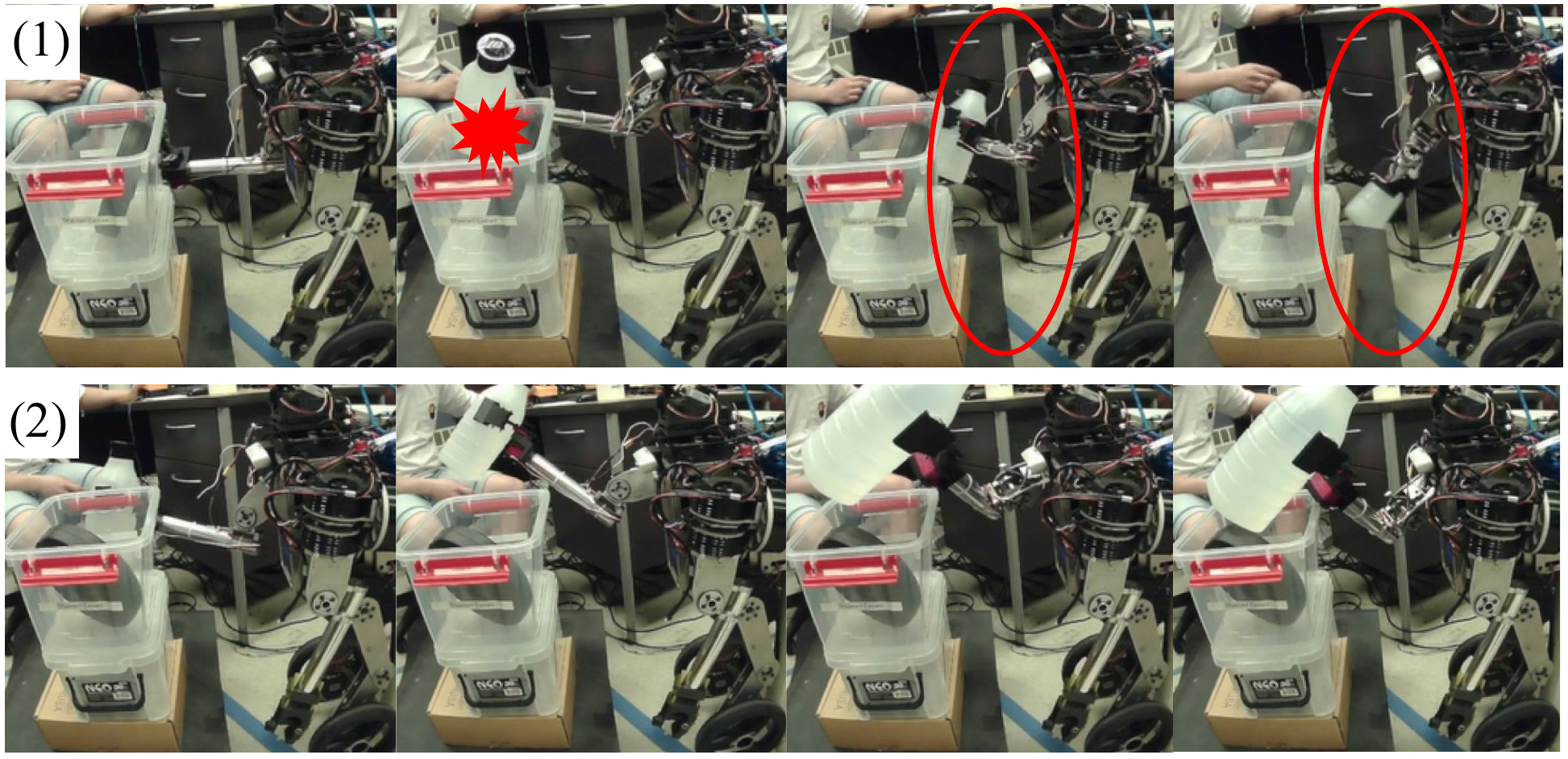}
    } 
    \subfloat[\label{fig:box_graph}]{
        \includegraphics[width=0.3\textwidth]{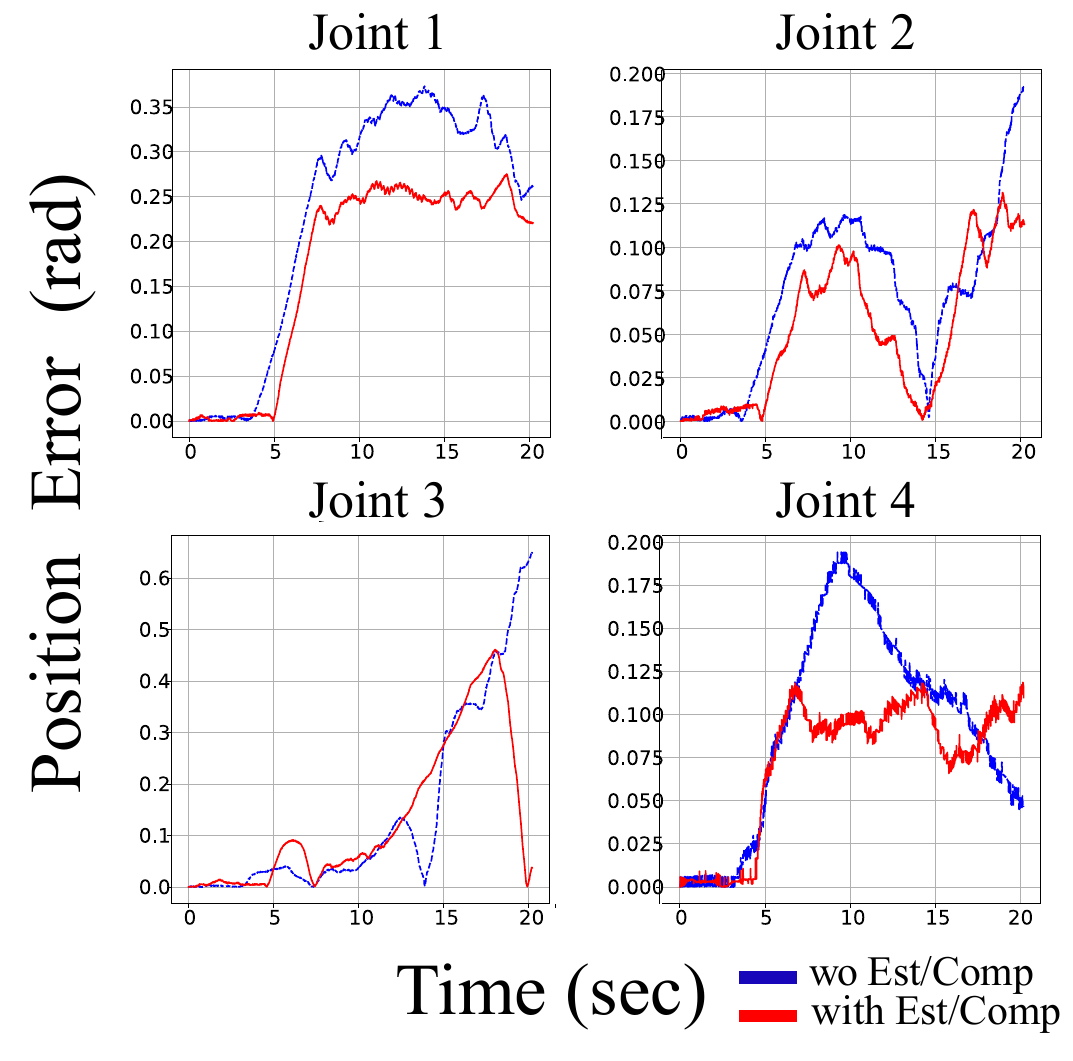}
    }

    \caption{Object pick-and-place task results. (a) Sequential frames from two trials: (1) without object dynamics compensation, resulting in instability and task failure (red highlight); and (2) with compensation using estimated inertial parameters, enabling successful task completion. (b) Mean absolute position error for each joint of the SATYRR arm over time, comparing the controller without inertial parameter estimation/compensation (wo Est/Comp) and the controller with estimation and compensation (with Est/Comp). The results show that using estimated parameters improves tracking accuracy. A slightly higher PD gain was used to ensure task success. }
    \label{fig:result_pick_and_place}
\end{figure*}

Figure~\ref{fig:result_pick_and_place} demonstrates that incorporating object dynamics compensation into a model-based controller via feedforward inverse dynamics significantly improves tracking performance and enables successful and safe task execution. Due to the use of low PD gains to maintain compliant behavior, which is essential for safe human-robot interaction, the manipulator lacked sufficient torque authority to lift the heavy object without compensation. By accounting for the object's inertial parameters, the controller was able to achieve accurate tracking while preserving compliance. As shown in Table~\ref{tab:mse_comparison}, consistent tracking improvements were observed across all joints, with the most significant gains in shoulder yaw (76.4\%) and elbow pitch (78.4\%).

\subsubsection{Whole-Body Bilateral Tele-Loco-Manipulation of Heavy Bottle on Hardware}
\label{VI-B2}
To evaluate the whole-body capabilities of the proposed approach, we validated the whole-body bilateral teleoperation framework through hardware experiments using SATYRR and the HMI across two different scenarios. Throughout the experiments, we investigate: (1) the impact of incorporating object inertial parameter estimation on heavy-object manipulation within the DMM teleoperation framework; (2) how the estimator improves the reliability of haptic force feedback; and (3) the feasibility of dynamic whole-body loco-manipulation involving lifting, moving, squatting, and releasing while handling a heavy object. \rev{The estimated object dynamics are incorporated into the controller once available, and in the current teleoperation setup, this update is naturally triggered by the operator during the pre-lifting stabilization phase.}

\input{table/manip_result_table}

In the first experiment, SATYRR lifts a 3.3\,kg water bottle (one-third of the robot's weight), moves backward, and returns to the original position to release the object via whole-body teleoperation. Three cases were compared: (i) \textbf{wo Compensation} — object dynamics are not compensated; the human manually responds to disturbances using visual and haptic feedback, leaning backward to maintain balance. 
(ii) \textbf{with Compensation wo Haptic} — object dynamics are automatically compensated, but the human receives no haptic feedback. 
(iii) \textbf{with Compensation \& Haptic} — object dynamics are automatically compensated, and the human receives haptic feedback.  

\begin{figure*}[t]
    \centering
    \subfloat[wo Compensation]{%
        \includegraphics[width=0.32\linewidth]{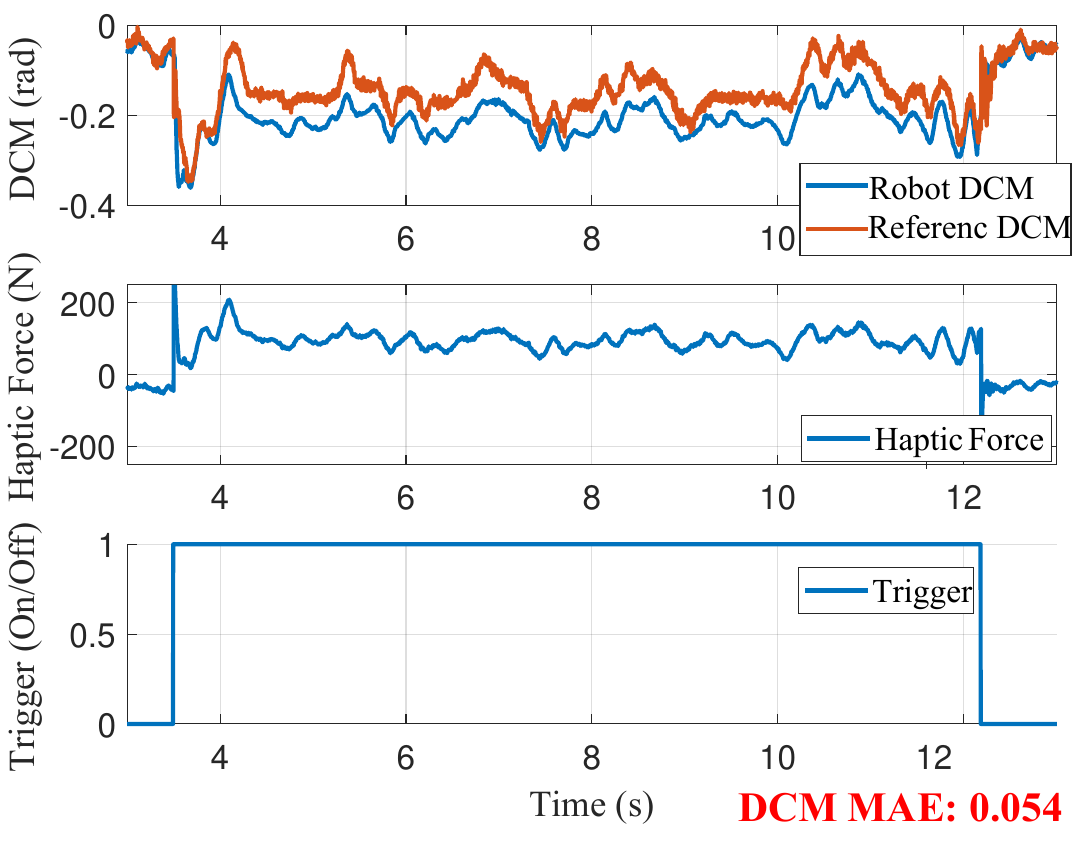}%
    }\hspace{0.01\linewidth}
    \subfloat[with Compensation wo Haptic]{%
        \includegraphics[width=0.32\linewidth]{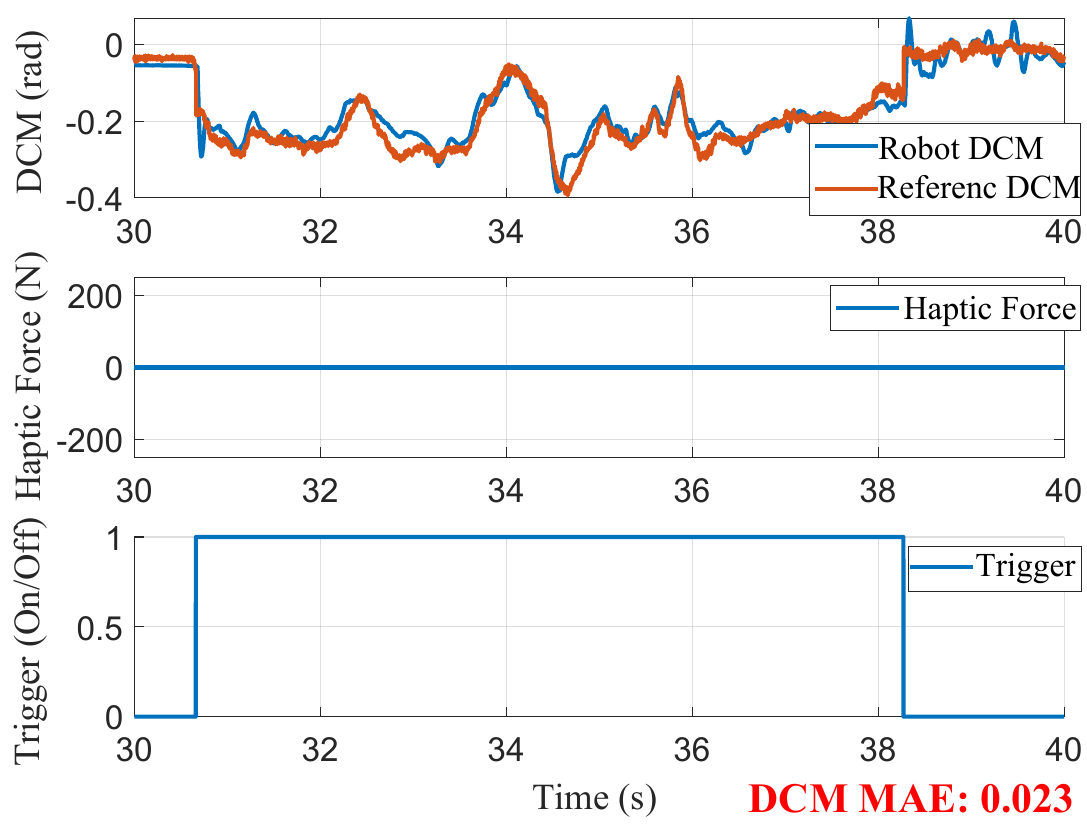}%
    }\hspace{0.01\linewidth}
    \subfloat[with Compensation \& Haptic]{%
        \includegraphics[width=0.32\linewidth]{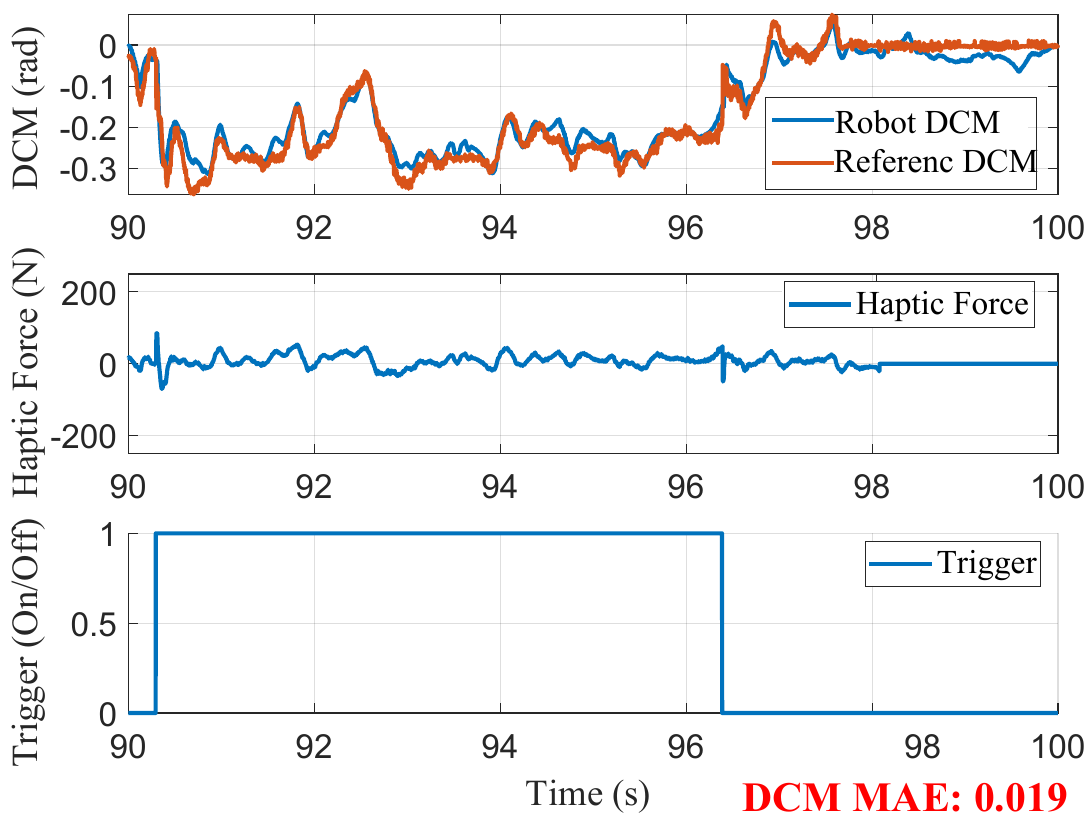}%
    }

    \caption{DCM tracking and haptic feedback during a whole-body loco-manipulation task (lifting, moving, and releasing a heavy water bottle). The robot’s desired and actual DCM are shown alongside the corresponding haptic forces and trigger signals, which indicate the onset of object lifting. Three experimental conditions are compared: (a) Without (wo) object dynamics compensation or haptic feedback, (b) With object dynamics compensation but no haptic feedback, and (c) With both object dynamics compensation and haptic feedback. Note that object parameter estimation enables successful task completion, whereas multiple attempts without estimation often resulted in failure. In addition, it enhances the effectiveness of haptic force feedback, allowing the human pilot to receive more informative cues and achieve better dynamic synchronization with the robot, as reflected in the improved DCM tracking performance (Mean Absolute Error).}
    \label{fig_dcm_obj_lift_comparison}
    \vspace{-0.5em}
\end{figure*}

Figure~\ref{fig_dcm_obj_lift_comparison} compares different methods in the whole body object lifting and delivery task. In the \textbf{wo Compensation} condition, the robot failed to complete the task in most trials (see attached video). One successful trial was recorded to analyze DCM tracking performance and evaluate the impact of haptic force feedback during teleoperation. Even in this case, significant DCM tracking errors and large haptic forces were observed.

\begin{figure*}[!t]
    \centering
    \subfloat[\footnotesize Snapshots of SATYRR lifting, moving, and squatting under bilateral teleoperation\label{fig:box_img}]{
        \includegraphics[width=0.9\textwidth]{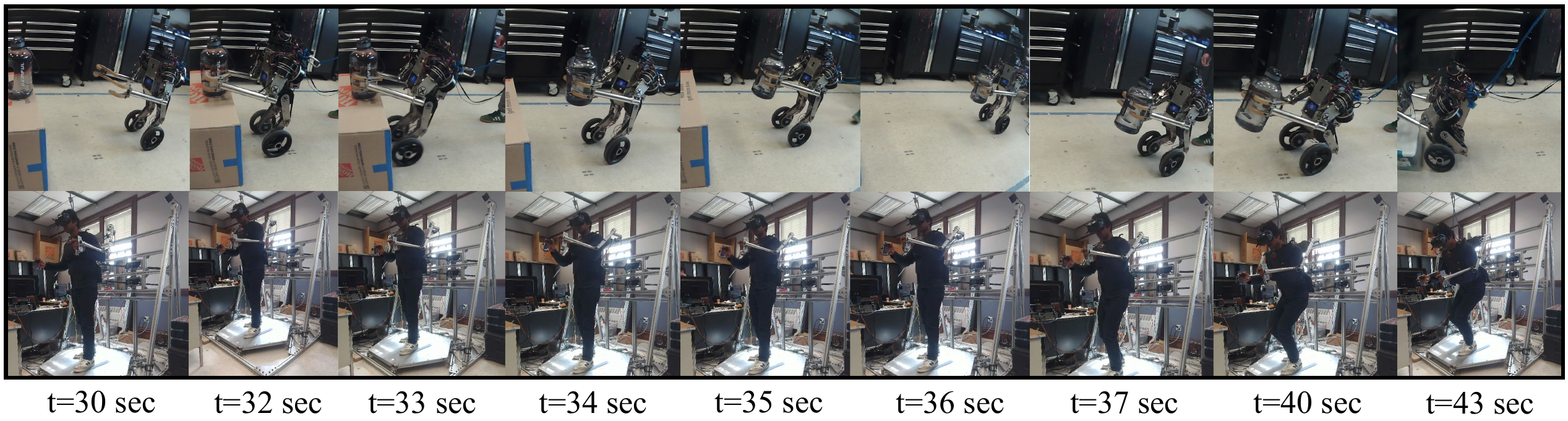}
    }\\[1ex]

    \subfloat[\footnotesize System response corresponding to the teleoperated task in (a)\label{fig:box_graph}]{
        \includegraphics[width=0.8\textwidth]{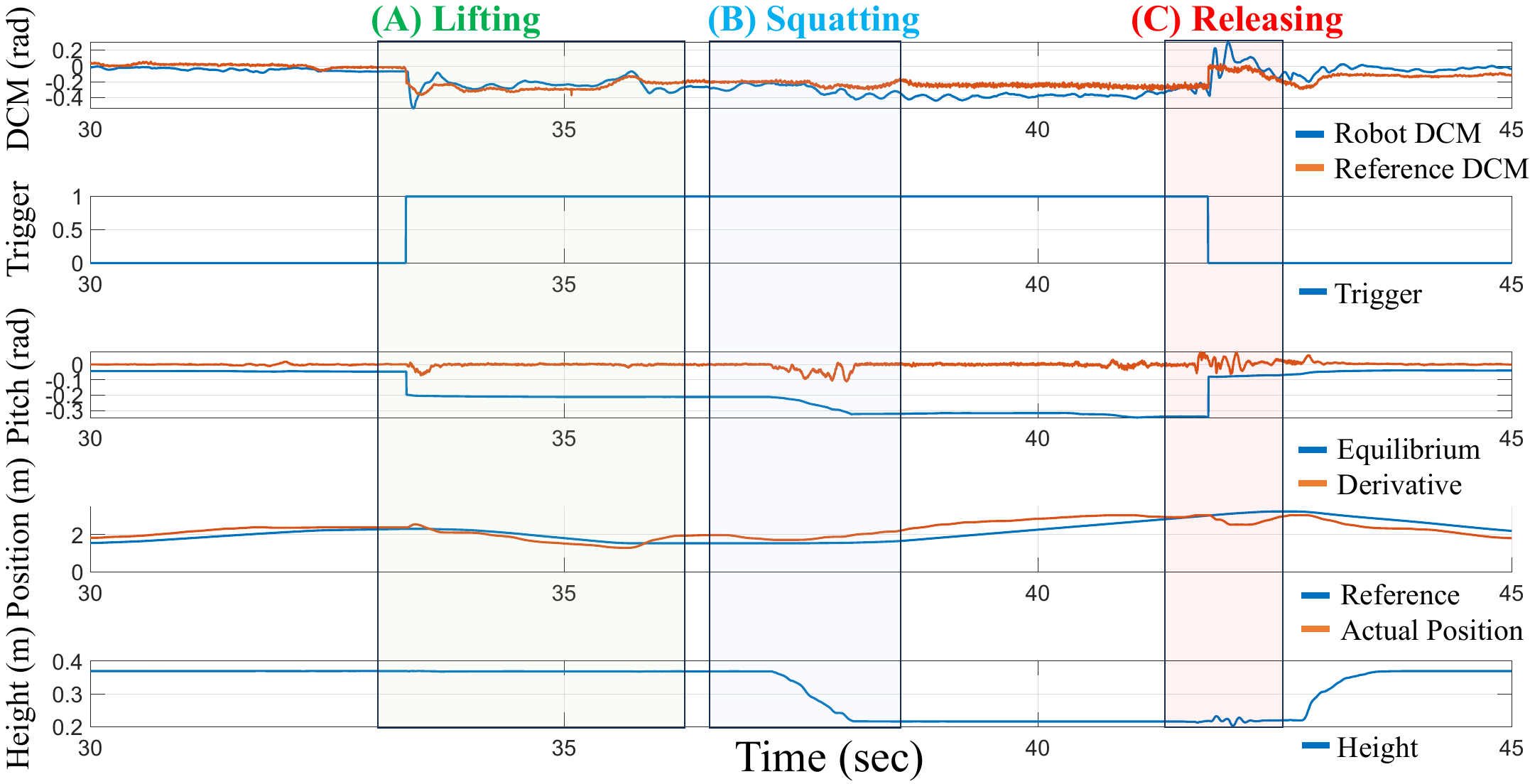}
    }
    
    \caption{
Results of whole-body tele-loco-manipulation. (a) Depicts the sequential execution of the task—lifting, moving, squatting, and releasing—guided by a human pilot. (b) Shows the corresponding dynamic responses during a trial without haptic feedback, DCM tracking, or active height modulation. The green-shaded region marks the activation of equilibrium adaptation, followed by the squatting phase (blue) and object release (red). (c) Presents a separate trial focusing on lifting and moving. During the green-shaded interval, online object parameter estimation is active, allowing the system to adapt to object dynamics. The haptic force remains relatively small and interpretable, enabling the human pilot to sense and utilize it effectively for decision-making. When estimation is turned off during the blue interval, the system begins to misinterpret object dynamics, leading to increased haptic force magnitudes. In the red-shaded region, where both estimation and adaptation are disabled, the haptic force becomes erratic and overwhelming, disrupting human decision-making and resulting in task failure.
}

\label{fig:locomanipulation}
\end{figure*}

Empirically, haptic feedback improves disturbance perception compared to vision alone, but increases physical fatigue due to continuous resistance. The proposed framework, integrating object parameter estimation and adaptive equilibrium control, improves DCM tracking while reducing both cognitive and physical load, allowing the pilot to focus on high-level manipulation as the robot compensates for unknown object dynamics. Combined with parameter estimation, haptic feedback provides more accurate force cues, enhancing interaction awareness, particularly when visual access to contact is limited \cite{purushottam2023dynamic, purushottam2024wheeled}.

In the second experiment, we evaluate a more challenging scenario by incorporating squatting during object delivery. Fig.~\ref{fig:locomanipulation}(a) shows the task execution, and (b) presents the corresponding DCM tracking and equilibrium evolution. The results demonstrate that the framework enables dynamic whole-body loco-manipulation, allowing the pilot to focus on high-level actions while low-level control (e.g., equilibrium adaptation) is handled autonomously, improving stability during challenging motions.

\rev{In the current implementation, VLM-based parameter initialization (2--3~s) is completed prior to interaction via the human-in-the-loop teleoperation protocol, where a brief stabilization phase naturally precedes lifting. If lifting begins before initialization, the controller operates with inaccurate parameters, degrading equilibrium estimation and increasing tracking error in the initial phase. Stability is maintained through feedback control, and performance improves once estimates are updated.}


\subsection{Collision-Avoidance Manipulation with CBF in Simulation}
\label{VI-B3}

\rev{We further investigate the benefits of integrating multi-stage object parameter estimation into safety-critical manipulation tasks, such as collision avoidance between the manipulated object and its surroundings. This extends the applicability of our method to safety-critical scenarios within a control-theoretic framework.}

\begin{figure}[t]
\centering
\begin{minipage}{\linewidth}
    \centering
    \includegraphics[width=0.8\linewidth]{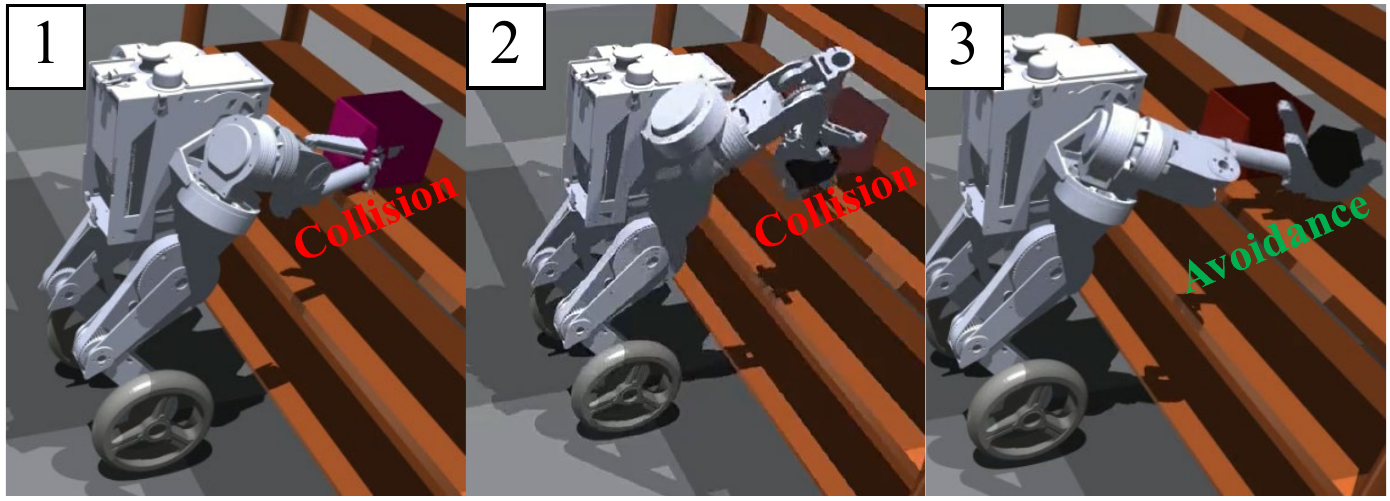}
\end{minipage}

\vspace{1em}  

\begin{minipage}{\linewidth}
    \centering
    \includegraphics[width=1\linewidth]{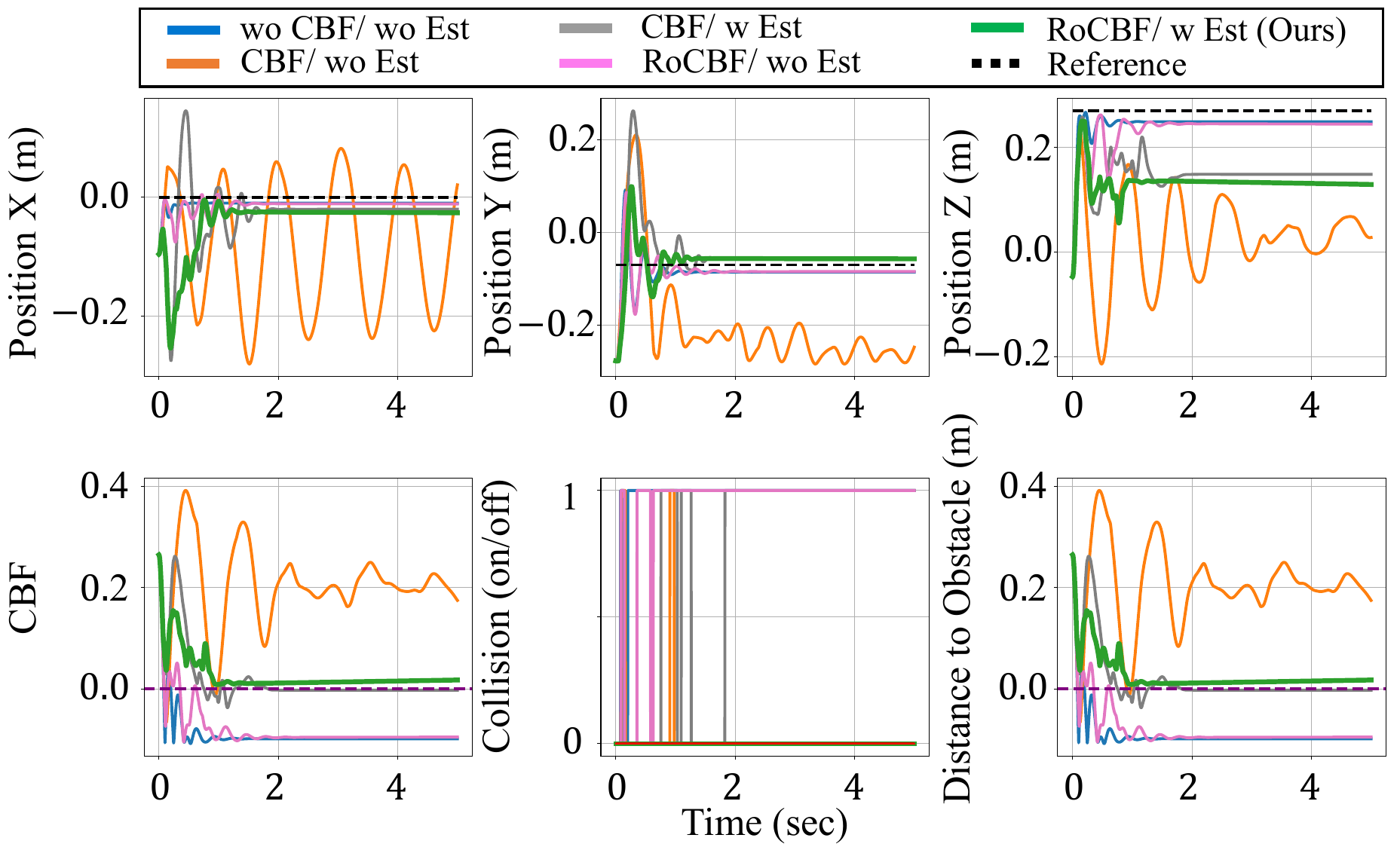}
\end{minipage}

\caption{Results of safety-critical manipulation control with object parameter estimation and control barrier function. The robot aims to avoid collision between a static obstacle (purple box) and a manipulated object (black) held in the right hand. Top: (1) Without CBF or estimation, the robot collides with the environment. (2) With CBF but without object parameter estimation, the robot still collides due to unmodeled object dynamics. (3) With both robust CBF and online object parameter estimation, the robot successfully avoids collision. Bottom: Task-space tracking (X, Y, Z), CBF value, collision flag, and distance to obstacle are plotted over time. Only the proposed method (RoCBF / w Est) satisfies CBF safety conditions and avoids collisions, demonstrating that estimation adapts dynamic constraints to the object, while the robust formulation ensures safety under estimation uncertainty.}
\label{fig_cbf}
\end{figure}

\rev{The experimental results are shown in Fig.~\ref{fig_cbf}. We observe that the CBF filter becomes significantly more effective when the object estimator is incorporated, suggesting that reducing dynamics uncertainty improves the reliability of constraint enforcement. In contrast, without accurate object dynamics, the controller may violate safety constraints due to model mismatch.}

\rev{We emphasize that these results are demonstrated in simulation. Extending this approach to hardware requires careful consideration of sim-to-real dynamics discrepancies, which we leave for future work.}

%% file: table/obj_shape_table.tex
\definecolor{darkgreen}{rgb}{0.0, 0.4, 0.0}
\begin{table}[t]
\centering
\caption{\rev{Estimation performance by object shape class (simulation). Accuracy degrades with deviation from cuboid geometry, especially for inertia and CoM. Best: green bold; second-best: bold.}}
\setlength{\tabcolsep}{3pt}
\renewcommand{\arraystretch}{1.05}
\begin{tabular}{lcccc}
\toprule
Shape class (78) & CI $\uparrow$ & Mass NMAE $\downarrow$ & Inertia NMAE $\downarrow$ & CoM [mm] $\downarrow$ \\
\midrule
Cuboid (9)   & \textbf{\textcolor{darkgreen}{0.725}} & \textbf{0.139} & \textbf{0.166} & \textbf{\textcolor{darkgreen}{1.6}} \\
Cylinder (8)  & 0.688 & 0.166 & 0.447 & 3.7 \\
Irregular (20) & 0.602 & 0.169 & 0.263 & 9.2 \\
Unknown (41)   & 0.684 & \textbf{\textcolor{darkgreen}{0.118}} & \textbf{\textcolor{darkgreen}{0.043}} & \textbf{\textcolor{darkgreen}{1.6}} \\
\midrule
All       & 0.665 & 0.142 & 0.173 & 4.1 \\
\bottomrule
\end{tabular}
\label{tab:cuboidness_shape}
\end{table}

%% file: table/manip_result_table.tex
\definecolor{darkgreen}{rgb}{0.0, 0.4, 0.0}

\begin{table}[t]
\centering
\caption{Mean squared joint position error (in radians) averaged over five objects during the pick-and-place task. Corresponds to Fig.~\ref{fig:result_pick_and_place}(b).}
\label{tab:mse_comparison}
\scriptsize
\setlength{\tabcolsep}{9pt} 
\renewcommand{\arraystretch}{1.1}

\begin{tabular}{lcccc}
\toprule
\textbf{MSE (5 cases)} & \makecell[c]{\textbf{Shoulder} \\ \textbf{Pitch}} & \makecell[c]{\textbf{Shoulder} \\ \textbf{Roll}} & \makecell[c]{\textbf{Shoulder} \\ \textbf{Yaw}} & \makecell[c]{\textbf{Elbow} \\ \textbf{Pitch}} \\
\midrule
WO Compensation & 0.0297 & 0.0128 & 0.1994 & 0.0139 \\
W Compensation  & 0.015  & 0.0097 & 0.047  & 0.003 \\
\textbf{Improvement} 
& \textbf{\textcolor{darkgreen}{49.5\%}} 
& \textbf{\textcolor{darkgreen}{24.2\%}} 
& \textbf{\textcolor{darkgreen}{76.4\%}} 
& \textbf{\textcolor{darkgreen}{78.4\%}} \\
\bottomrule
\end{tabular}
\end{table}

%% file: Conclusions_FutureWork.tex
\section{Conclusion and Future Work}
\label{Conclusion_and_FutureWork}

This work introduces a whole-body bilateral teleoperation framework with multi-stage online estimation of an object's inertial parameters for a wheeled humanoid robot. We specifically address tasks involving lifting, transporting, and releasing relatively heavy objects (e.g., water bottles or liquid-filled containers), which demands whole-body coordination, awareness of the object's physical properties, and rapid adaptability to remain robust against external disturbances. Incorporating multi-stage object parameter estimation improves manipulation tracking performance while maintaining compliant behavior. It also enables the human operator to focus on locomotion and manipulation by automatically adjusting the system's equilibrium point to account for the object's dynamics, and enhances haptic force feedback for synchronized human–robot interaction. The proposed framework is validated in simulation and custom hardware, demonstrating its effectiveness for dynamic whole-body teleoperation tasks.

\rev{
Despite these results, several limitations remain. First, the cuboid-based geometric assumption introduces modeling errors for objects with complex or non-uniform geometry, particularly affecting CoM and inertia estimation. Second, the current system relies on VLM-based priors, which introduce a few-second latency that may limit responsiveness in rapid interaction scenarios. Finally, the framework depends on human-in-the-loop coordination and does not yet support fully autonomous operation.
}

\rev{
Future work will address these limitations by improving geometry representations for more accurate inertial modeling and reducing perception latency. We further aim to integrate the estimation framework with data-driven control, including predictive (e.g., MPC) and learning-based approaches, to enable more anticipatory and adaptive loco-manipulation. Finally, we plan to extend the framework to fully autonomous settings and validate safety-critical control (e.g., CBF) on hardware.
}

%% file: appendix.tex
\section{Appendix}
\label{Appendix}

\begin{figure}[t]
\centering
\includegraphics[width=\linewidth]{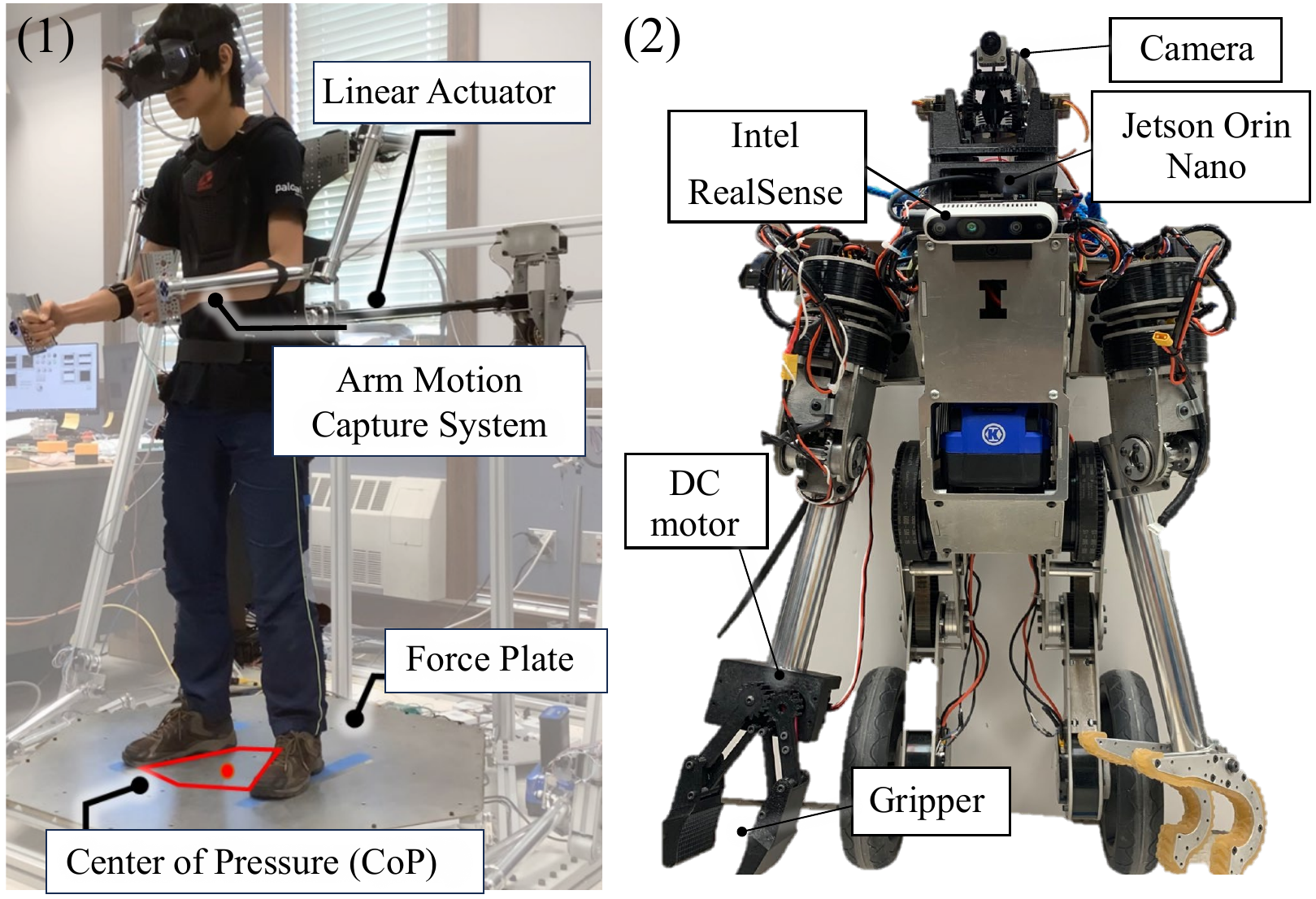}
\caption{Whole-body bilateral teleoperation system comprising a human–machine interface (HMI) and the wheeled humanoid SATYRR. (1) The HMI captures human kinematic and dynamic information using limb motion capture system, linear actuators, and a force plate, while providing haptic force feedback. (2) SATYRR is a wheeled humanoid with two wheels, articulated legs, arms, and a torso. A camera mounted on top provides the human pilot with the robot’s visual perspective. The gripper is actuated by a DC motor controlled via an Arduino Nano, while object size estimation is performed using an Intel RealSense camera and a Jetson Orin Nano.}

\label{fig3_system_description}
\end{figure}

\subsection{Hardware and Implementation Details}
\label{appendix::system} 

\subsubsection{SATYRR Hardware Setup}
SATYRR is a wheeled humanoid robot with anthropomorphic legs and two 4-DoF arms, where each leg employs an active wheel instead of a foot (Fig.~\ref{fig3_system_description}). The lower body is driven by four quasi-direct drive motors~\cite{Katz_thesis}, each current-controlled and connected via CAN to an NI sbRIO-9626 controller. Each motor includes a 12-bit encoder, and a torso-mounted VN-100 IMU provides orientation estimates at up to 200~Hz. The system is powered by a 24V, 4Ah battery, weighs 12~kg, has a nominal hip height of 0.4~m, and operates at 400~Hz. Communication with the human–machine interface (HMI) is performed via UDP. A custom 1-DoF gripper is actuated by a DC motor with PWM control.

\subsubsection{Human Machine Interface Hardware Setup}
The HMI system~\cite{wang2021dynamic} consists of a high-force haptic device, a force plate, and a passive MoCap exoskeleton. The haptic device uses four quasi-direct drive motors on passive gimbals, capable of delivering up to 100~N per axis to the operator’s torso while measuring position and applying multi-directional spring-like forces. The force plate integrates six uniaxial load cells to estimate ground reaction forces and center of pressure (CoP). The system is controlled by an NI cRIO-9082 real-time controller running at 1~kHz.

\subsubsection{Implementation Details}
The system is distributed across multiple computing units and threads. The main controller (motor control, state estimation, locomotion mapping, and manipulation) runs on an onboard sbRIO at $\sim$400~Hz. The object parameter estimation module, based on IsaacGym, runs on a separate desktop and communicates with the robot controller via UDP. Object size estimation, VLM inference, and gripper control are executed on an NVIDIA Jetson Nano (see Fig.~\ref{fig3_system_description}).

\rev{The Stage-3 DH-CEM refinement employs 500 samples per iteration with 50 elites (10\%) over 3 iterations. Mass and inertia parameters are sampled with a standard deviation of 10\% of the current mean, while the center of mass is sampled with a fixed standard deviation of 0.01. Shape parameters follow a coarse-to-fine strategy, with larger initial uncertainty and reduced perturbation ($\sim$2\% of object size) in later iterations. Multi-hypothesis initialization is derived from the vision/VLM prior, improving robustness to prior uncertainty.}

\subsection{High-Fidelity Simulation via Sim-to-Real Adaptation}
\label{appendix::sim-to-real}

\subsubsection{Sim-to-Real Gap.}
\label{appendix::sim_to_real_gap}
The \textit{reality gap} arises from discrepancies between simulation and real-world dynamics, where identical control inputs produce different system responses due to modeling errors, latency, and unmodeled effects \cite{zhao2020sim}. This discrepancy at the acceleration level can be expressed as:
\begin{equation}
\delta_{\ddot{q},t}^{\text{R}} := \ddot{q}_{\text{real},t} - \ddot{q}_{\text{sim},t} = 
\underbrace{\Delta f_{\mathrm{FD}}(\hat{x}_t, \hat{u}_t)}_{\text{Parametric Error}} + 
\underbrace{\delta_{\mathrm{np}}^{\text{R}}(\hat{x}_t, \hat{u}_t)}_{\text{Non-Parametric Error}},
\end{equation}
where \( f_{\mathrm{FD}}(\cdot) \) denotes the forward dynamics:
\begin{equation}
f_{\mathrm{FD}}(\mathbf{q}, \dot{\mathbf{q}}, \boldsymbol{\tau}; \boldsymbol{\Phi}) = 
\mathbf{M}^{-1}(\mathbf{q}, \boldsymbol{\Phi}) 
\left( \boldsymbol{\tau} + \mathbf{J}_c^\top(\mathbf{q})\mathbf{f}_c - \mathbf{b}(\mathbf{q}, \dot{\mathbf{q}}, \boldsymbol{\Phi}) \right).
\end{equation}

Here, \( \boldsymbol{\Phi}_{\mathrm{sim}} \) and \( \boldsymbol{\Phi}_{\mathrm{real}} \) denote simulated and true inertial parameters. The non-parametric term \( \delta_{\mathrm{np}}^{\text{R}} \) captures effects such as friction nonlinearities, actuator dynamics, compliance, and latency. Under rigid grasp assumptions, we minimize contact-induced errors and primarily address parametric discrepancies.\\

\subsubsection{Sim-to-Real Adaptation.}
\label{appendix::sim_to_real_adapt} 
When a robot rigidly grasps an object, its proprioceptive signals encode the coupled robot--object dynamics. Our sampling-based estimator uses only robot's proprioception, making accurate modeling of object effects essential. We employ a high-fidelity simulation to minimize the reality gap and isolate the object's dynamic contribution, ensuring \( f_{\mathrm{ID}}(\cdot; \boldsymbol{\Phi}^{\text{R}}) \) replicates real-world trajectories under identical inputs.

To reduce the \textit{reality gap}, we perform offline system identification by tuning a subset of the robot’s model parameters \( \boldsymbol{\zeta} \in \mathbb{R}^9 \) in simulation, including joint friction (\( \mathbb{R}^4 \)), time delay (\( \mathbb{R}^1 \)), and control gains (\( \mathbb{R}^4 \)). These parameters, a subset of the full vector \( \boldsymbol{\Phi} \) defining the robot’s dynamics, are empirically identified as the primary contributors to the gap. Adjusting them sufficiently reduces discrepancies at the position and velocity levels under the assumptions that (1) contact dynamics are negligible and (2) both the object and articulated system behave as near-rigid bodies.

Following~\cite{baek2024online}, identification minimizes the discrepancy between simulated trajectories \( \mathbf{X}_S \) and real-world trajectories \( \mathbf{X}_T \), where each sample \( \mathbf{X} \in \mathbb{R}^8 \) contains joint positions \( \mathbf{q} \) and velocities \( \dot{\mathbf{q}} \) for the four joints of the SATYRR manipulator:
\begin{equation}
\mathcal{L}(\boldsymbol{\zeta}) = \frac{1}{m} \sum_{i=1}^{m} \sum_{t=1}^{T} d(\mathbf{X}_S^t, \mathbf{X}_T^t),
\label{costfunc}
\end{equation}
where \( d(\cdot,\cdot) \) is the mean squared error (MSE)~\cite{wensing2017linear}. We solve for \( \boldsymbol{\zeta} \) using particle swarm optimization (PSO) for its global search capability and rapid convergence. For real-time DH-CEM execution, we use the built-in PD controller, which is faster than torque-based control mode that requires costly sub-steps for accurate integration. \\

\subsection{Sim-to-Real Adaptation Evaluation}
\label{results:simtoreal}

Although domain randomization can mitigate the reality gap in reinforcement learning, it often leads to conservative motions, requires extensive tuning, and fails to align simulation and hardware trajectories~\cite{muratore2022robot, chen2020underactuated}. Despite recent improvements~\cite{he2025asap}, trajectory mismatches persist, degrading estimation accuracy due to limited feedback integration~\cite{baek2024toward, baek2024online, chen2020underactuated}. In contrast, this work explicitly aligns simulated and real trajectories, improving estimation via a sampling-based method in rigid-body simulation (Fig.~\ref{fig9}).

\begin{figure}[t]
    \centering
    \includegraphics[width=1\linewidth]{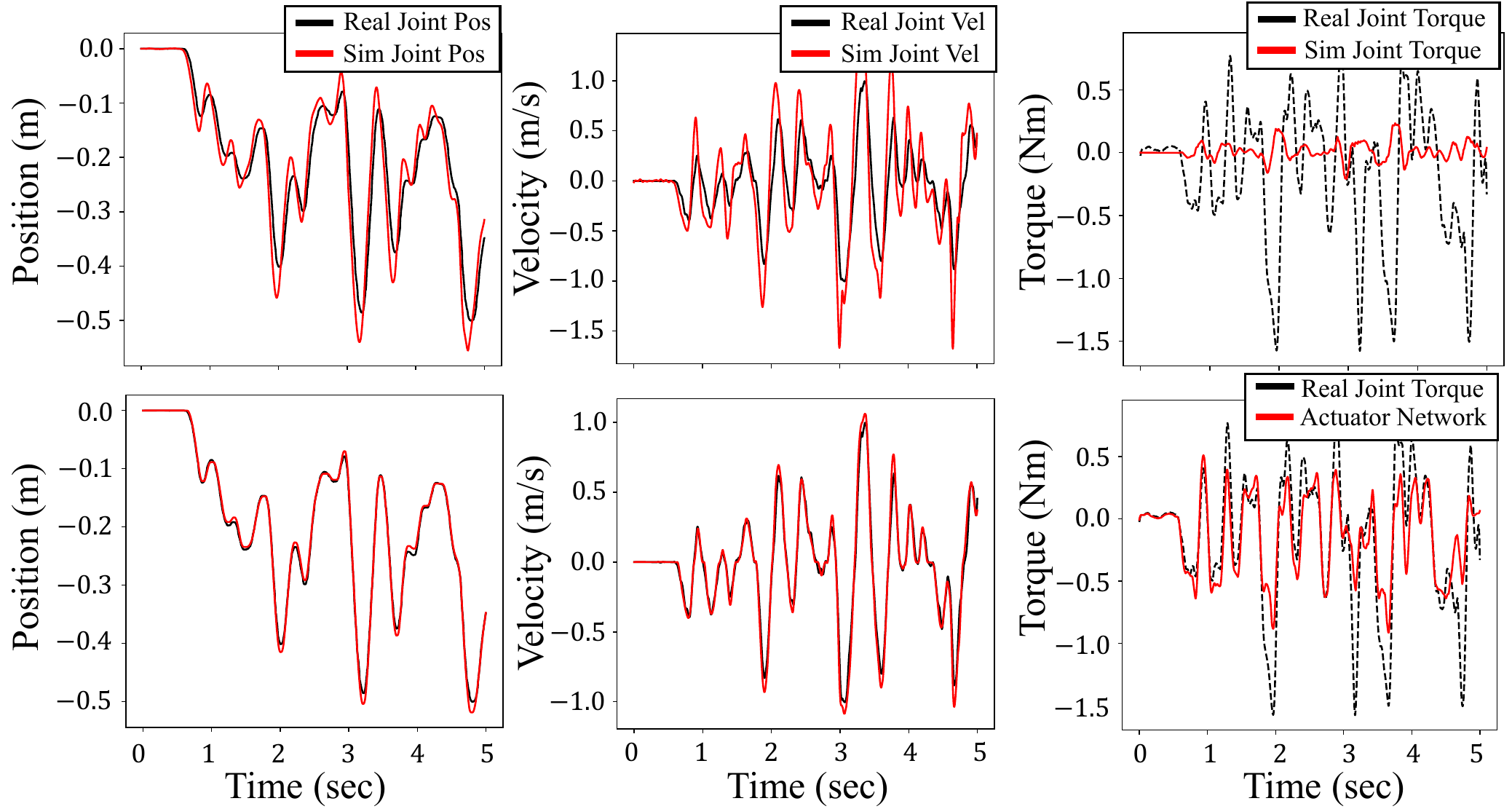}
    
    \caption{Comparison of joint-level trajectories between simulation and hardware to evaluate sim-to-real adaptation. The proposed method significantly reduces the \textit{reality gap} in position and velocity. While ActuatorNet further improves torque prediction, noticeable discrepancies remain, indicating limitations in fully capturing system dynamics.}

    \label{sim2real_fig}
\end{figure}

We evaluate three simulation settings: (i) \textbf{Baseline}, using identical controller gain, control frequency, zero friction, and CAD-based inertial parameters as hardware; (ii) \textbf{Ours}, incorporating identified control frequency (accounting for latency), estimated joint friction, and the same controller gain; and (iii) \textbf{Actuator Network}~\cite{hwangbo2019learning}, trained on SATYRR joint-level data to predict torque:
\begin{equation}
    \tau_j = h_{\theta}(q^*_j, q_j, \dot{q}^*_j, \dot{q}_j, K_p, K_d), \quad j = 1, 2, 3, 4,
\end{equation}
where \(q^*_j, \dot{q}^*_j\) denote desired states, \(q_j, \dot{q}_j\) measured states, and \(K_p, K_d\) PD gains. As described in Section~\ref{appendix::sim_to_real_adapt}, nine parameters are identified via particle swarm optimization (PSO) to minimize trajectory discrepancies between simulation and hardware.

Figure~\ref{sim2real_fig} shows that incorporating identified joint friction and latency significantly reduces the sim-to-real gap. We attribute the remaining discrepancy to two main factors: (i) latency, which delays sensing and actuation and alters effective time steps and velocity estimates; and (ii) friction, which resists low-speed motion, delaying position peaks and damping velocity reversals, consistent with hardware behavior.

Despite this, significant torque errors remain, primarily due to simplified dynamics, unmodeled effects (e.g., backlash, hysteresis), and hardware torque estimation, which relies on motor current and torque constants sensitive to temperature and manufacturing tolerances. The built-in PD controller partially mitigates these discrepancies through feedback and improves computational efficiency over custom controllers requiring more solver iterations. While the Actuator Network further reduces error, a residual gap remains, indicating that accurate torque alignment requires both improved simulation fidelity and more generalizable non-parametric models

\rev{\subsection{Excitation Sensitivity Analysis}
\label{appendix::excitation}
We analyze sensitivity to trajectory excitation by evaluating estimation performance across motions with varying excitation levels. Excitation is quantified using the log-determinant of the Fisher Information Matrix (FIM), which measures the overall information content of a trajectory; higher values indicate richer excitation across parameter directions.}

\begin{figure}[t]
\centering
\includegraphics[width=1\linewidth]{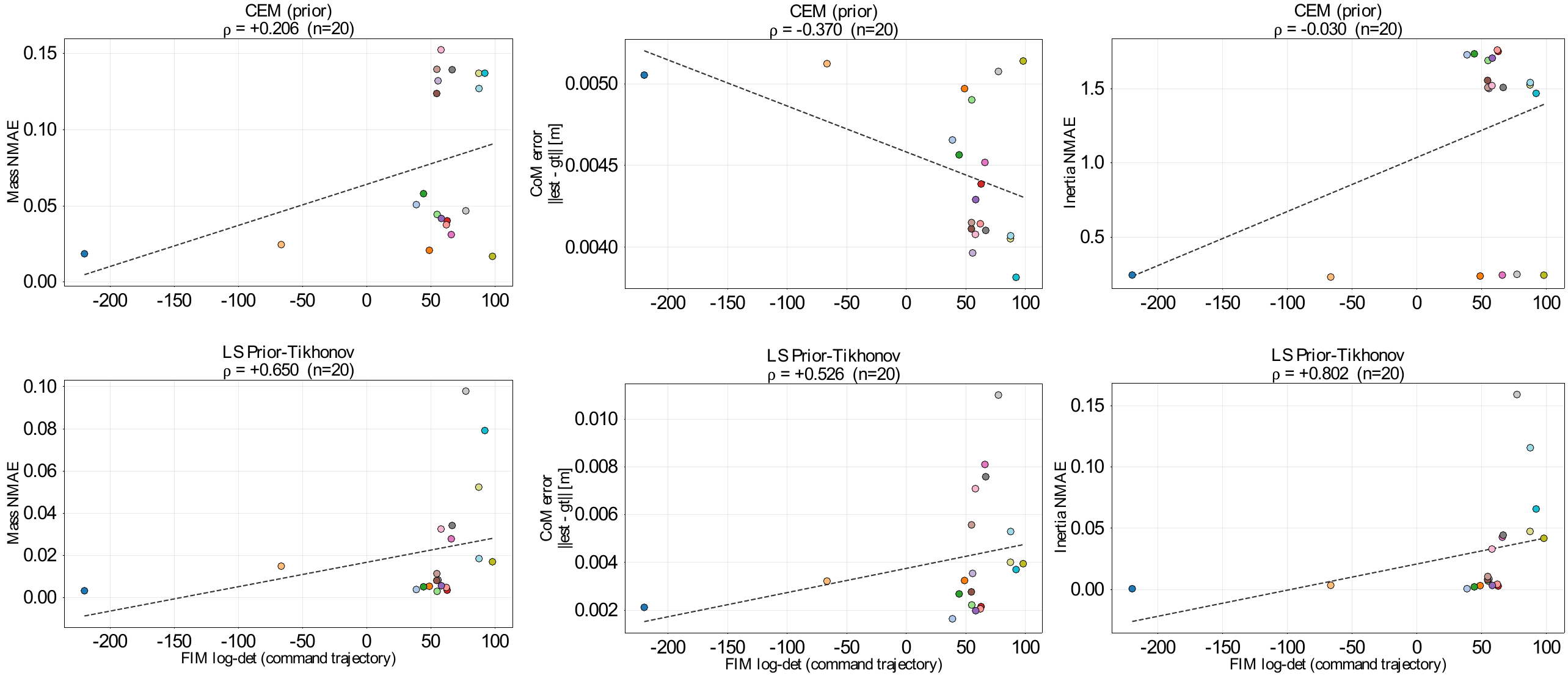}
\caption{\rev{Excitation sensitivity analysis comparing the proposed method (DH-CEM with prior) and a classical estimator with equivalent prior (LS Prior-Tikhonov). The x-axis shows the FIM log-determinant of the command trajectory and the y-axis shows estimation error. While LS Prior-Tikhonov exhibits strong dependence on excitation quality (e.g., $\rho = +0.802$ for 
inertia), the proposed method remains largely insensitive 
(e.g., $\rho = -0.030$ for inertia), demonstrating that the 
structural decomposition reduces excitation sensitivity without requiring dedicated exciting trajectories.}}
\label{fig:excitation}
\end{figure}

\rev{Fig.~\ref{fig:excitation} compares the proposed method (DH-CEM with prior) with a classical least-squares estimator (LS Prior-Tikhonov) for mass, center of mass, and inertia estimation. As shown in Fig.~\ref{fig:excitation}, least-squares exhibits strong dependence on excitation quality ($\rho \approx 0.80$ for inertia), whereas the proposed method shows significantly reduced sensitivity ($\rho \approx -0.03$).}

\rev{This difference arises from the estimation structure. Classical least-squares directly infers inertial parameters from observed motion, making it highly sensitive to poorly excited directions. In contrast, the proposed method leverages geometric and semantic priors and a decoupled formulation, where inertia is derived from estimated mass and geometry rather than directly identified from interaction data. This reduces the impact of weakly observable directions and improves robustness under limited excitation.}

\rev{Nevertheless, degenerate motions (e.g., near-static or single-axis trajectories) remain insufficient for reliable estimation, while natural task-driven motions (e.g., lifting) provide adequate excitation in practice.
}

\begin{figure}[t]
\centering
\includegraphics[width=0.8\linewidth]{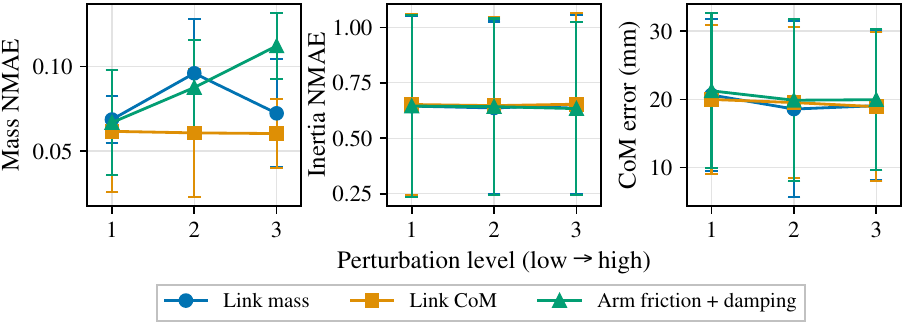} 
\caption{\rev{
Effect of residual robot model bias on object parameter estimation. 
Perturbation level (1–3) denotes increasing bias magnitude within each bias type: 
link mass (+5/10/20\%), link CoM (+5/10/20 mm), and joint friction/damping (+10/30/50\%). 
Results are averaged over 9 objects. Object mass shows the highest sensitivity to model mismatch, 
while inertia and CoM remain relatively stable.
}}
\label{fig:robot_bias_paper_1x3}
\vspace{-0.5cm}
\end{figure}

\rev{\subsection{Robustness to Robot Model Bias}}
\label{appendix::robot_bias}
\rev{
We evaluate the robustness of the proposed method to residual robot model mismatch by injecting controlled bias into the simulation while keeping ground-truth object parameters fixed. Three types of bias are considered: link mass scaling (+5/10/20\%), link CoM offset (+5/10/20 mm), and joint friction/damping increase (+10/30/50\%).}
\rev{
Fig.~\ref{fig:robot_bias_paper_1x3} shows that estimation error increases gradually with bias magnitude but remains bounded. Object mass is the most sensitive to mismatch, suggesting that mass estimation partially compensates for unmodeled robot dynamics. In contrast, inertia and CoM remain relatively stable, as they are primarily constrained by geometric priors and the decoupled estimation structure.}
\rev{
These results indicate that robot model bias propagates into object parameter estimation in a structured manner, rather than causing arbitrary degradation, demonstrating robustness to moderate mismatch.
}

%% file: biographies.tex
\begin{IEEEbiography}
[{\includegraphics[width=1in,height=1.25in,clip,keepaspectratio]
{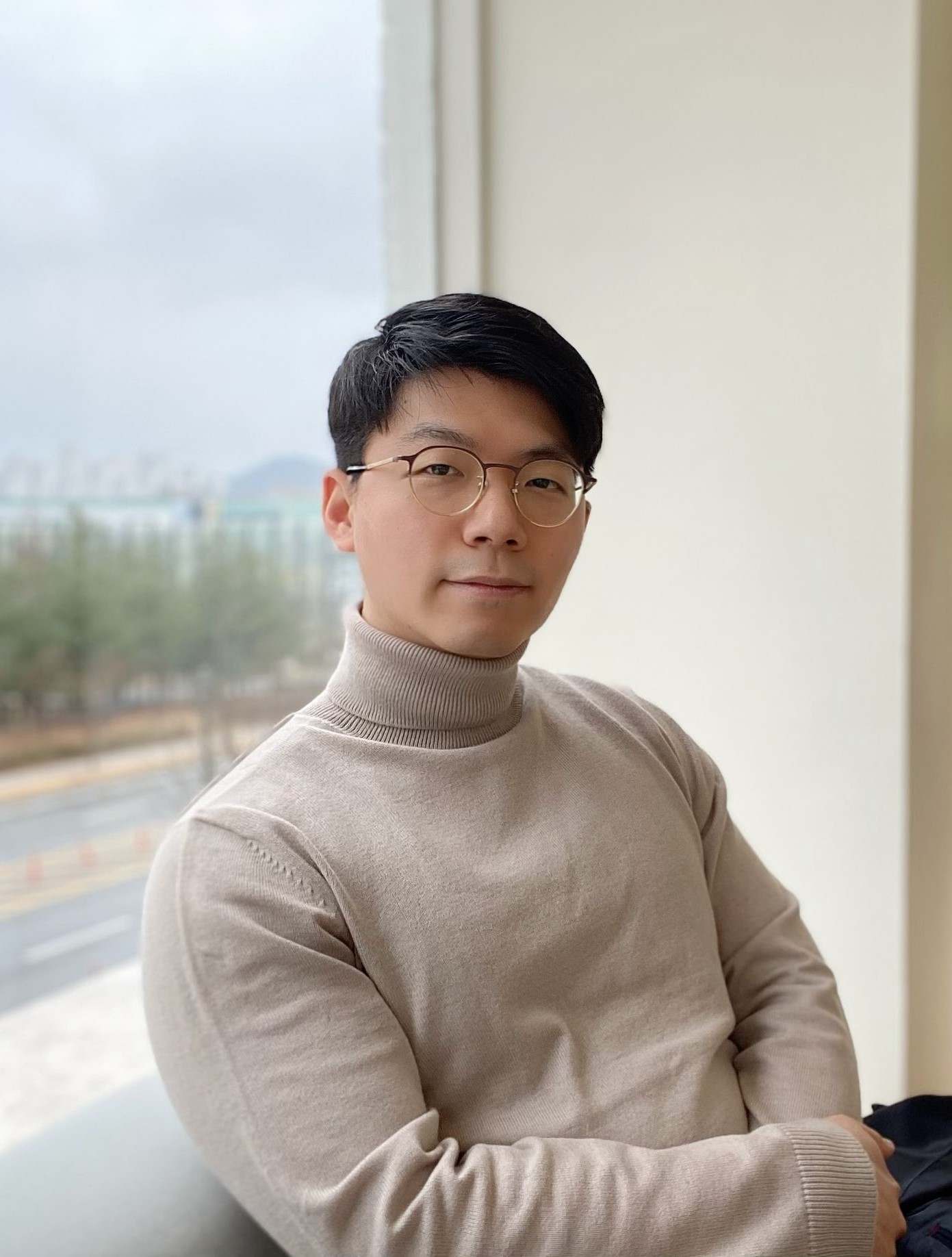}}]
{Donghoon Baek}
is a Postdoctoral Fellow with the School of Interactive
Computing, Georgia Institute of Technology. He received the Ph.D. degree in mechanical science and engineering from the University of Illinois Urbana-Champaign in 2025. He received the M.S. degree in robotics from the Korea Advanced Institute
of Science and Technology (KAIST) in 2019,
and the B.S. degree in robotics from Kwangwoon University in 2017. His research interests include control, robotics, and robot learning.
\end{IEEEbiography}

\begin{IEEEbiography}
[{\includegraphics[width=1in,height=1.25in,clip,keepaspectratio]
{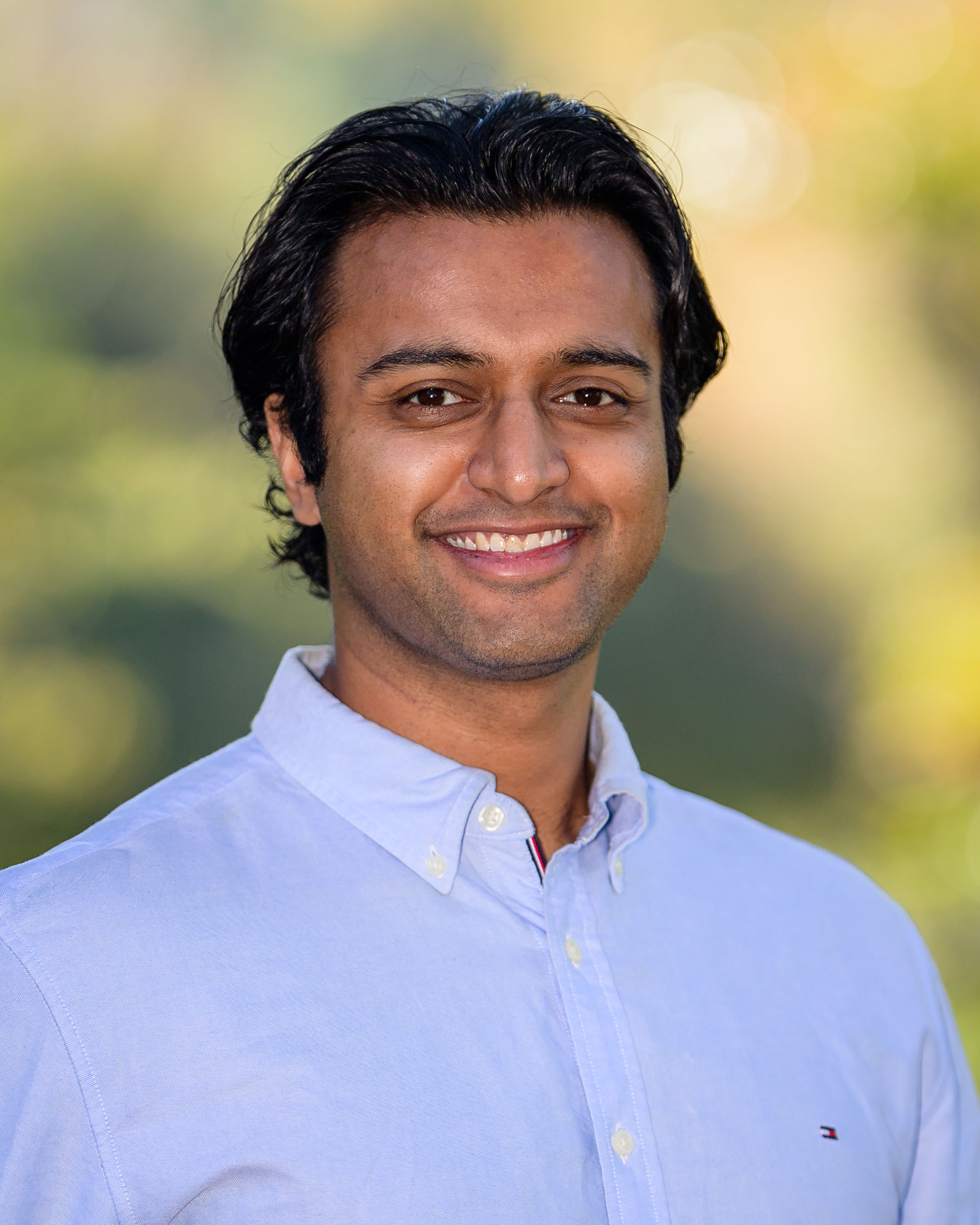}}]
{Amartya Purushottam}
received the B.S., M.S., and Ph.D. degrees in electrical and computer engineering from the University of Illinois Urbana-Champaign in 2018, 2021, and 2025, respectively. His research focuses on teleoperation for humanoid robotic systems, spanning whole-body control, haptic feedback, and dexterous manipulation.
\end{IEEEbiography}

\begin{IEEEbiography}
[{\includegraphics[width=1in,height=1.25in,clip,keepaspectratio]
{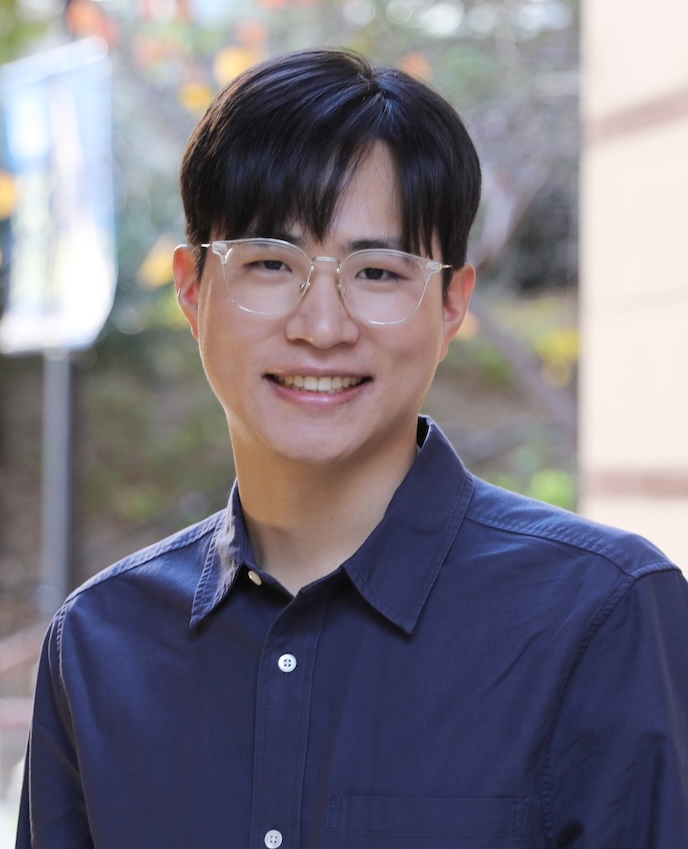}}]
{Jason Jangho Choi}
is an assistant professor in the Electrical and Computer Engineering Department at the University of California, Los Angeles. He received the Ph.D. degree in mechanical engineering at UC Berkeley in 2025. He received the B.S. degree in mechanical engineering from Seoul National University in 2019. His research focuses on safe and multi-agent decision-making for next-generation robots and autonomous systems. More broadly, his work lies at the intersection of control, robotics, and learning.
\end{IEEEbiography}

\begin{IEEEbiography}
[{\includegraphics[width=1in,height=1.25in,clip,keepaspectratio]
{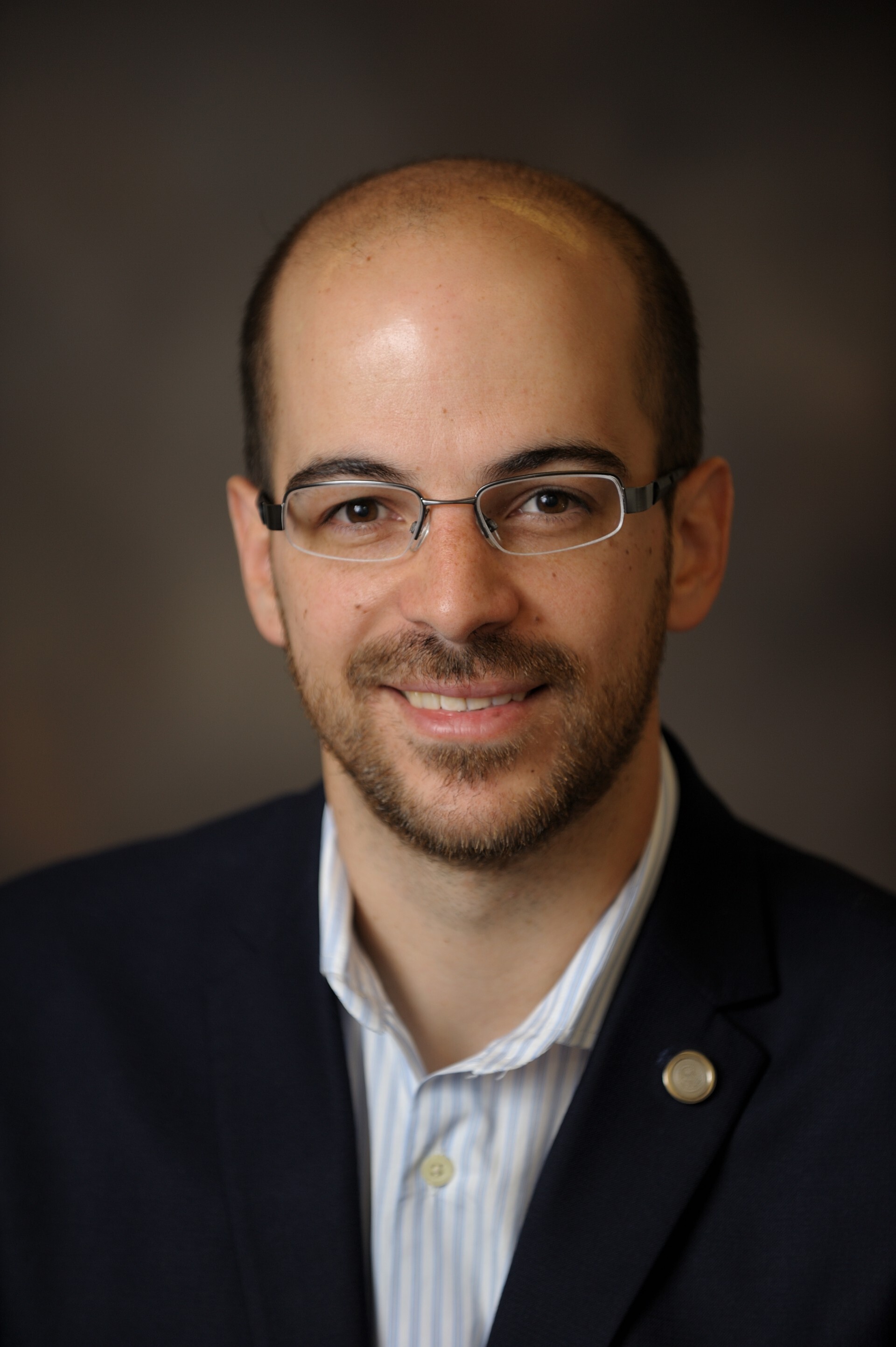}}]
{Joao Ramos} is a Research Scientist at Skild AI. Previously, he served as a Senior Technical Lead at Boston Dynamics, developing actuation, sensing, and structural design approaches that enabled the new generation of electric Atlas (both prototype and production versions). From 2019 to 2024, Dr. Ramos was an Assistant Professor at the University of Illinois at Urbana-Champaign (UIUC) and director of the RoboDesign Lab, where he earned the 2021 NSF CAREER Award. Before joining UIUC, he completed his PhD in Mechanical Engineering (2018) and a postdoctoral fellowship at MIT's Biomimetic Robotics Laboratory. His research focuses on the design and control of dynamic robotic systems, legged locomotion dynamics, actuation systems, and human-machine interfaces.
\end{IEEEbiography}